\documentclass[11pt]{article}

\usepackage{psfrag,amsmath,amsbsy,amsfonts,amssymb,amsthm,dsfont,fullpage,units}
\usepackage{graphicx}
\usepackage{algorithm,algorithmic,mathtools}
\usepackage{color,cases}
\usepackage{tikz}
\usetikzlibrary{calc,shapes}
\usepackage{subfig}
\usepackage{hyperref}

\newtheorem{assumption}{Assumption}
\newtheorem{lemma}{Lemma}[section]
\newtheorem{proposition}{Proposition}[section]
\newtheorem{theorem}{Theorem}[section]
\theoremstyle{definition}

\newtheorem{definition}{Definition}

\DeclareMathOperator{\dict}{dict}

\DeclareMathOperator{\corr}{corr}
\newcommand{\citep}{\cite}
\newcommand{\citet}{\cite}

\newcommand\E{\mathbb{E}}
\newcommand\R{\mathbb{R}}

\def\Pbb{\mathbb{P}}

\DeclareMathOperator{\uniqint}{Uniq-intersect}
\DeclareMathOperator{\coeff}{coeff}

\newcommand\sign{\operatorname{sign}}
\newcommand{\subsamp}{\overline{n}}

\newcommand{\ainit}{\ensuremath{\bar{a}}}

\newcommand{\Ainit}{\ensuremath{\bar{A}}}

\newcommand{\spindex}{\ensuremath{s}}

\newcommand{\errA}{\ensuremath{\epsilon_A}}

\newcommand{\singmin}{\ensuremath{\sigma_{\min}}}
\newcommand{\singmax}{\ensuremath{\sigma_{\max}}}

\newcommand{\hL}{\ensuremath{\widehat{L}}}

\newcommand{\expec}[1]{\mathbb{E}\left[#1\right]}
\newcommand{\prob}[1]{\mathbb{P}\left[#1\right]}
\newcommand{\twonorm}[1]{\left\| {#1} \right\|_2}
\newcommand{\iprod}[2]{\langle #1, #2 \rangle}
\newcommand{\frob}[1]{\left\|#1\right\|_F}
\newcommand{\set}[1]{\left\{#1\right\}}
\newcommand{\order}{\ensuremath{\mathcal{O}}}

\makeatletter
\newcommand{\customlabel}[2]{%
   \protected@write \@auxout {}{\string \newlabel {#1}{{#2}{\thepage}{#2}{#1}{}} }%
   \hypertarget{#1}{#2}
}
\makeatother

\newcommand\inner[1]{\ensuremath{\langle #1 \rangle}}

  \DeclareMathOperator{\nbd}{\mathcal{N}}

\DeclareMathOperator{\supp}{Supp}

\def\tha{{\mbox{\tiny th}}}

\DeclarePairedDelimiter\norm{\lVert}{\rVert}

\DeclarePairedDelimiter\onenorm{\lVert}{\rVert_1}

 \DeclarePairedDelimiter\abs{\lvert}{\rvert}

 \def\0{{\bf 0}}

\def\viz{{viz.,\ \/}}

\def\qed{\hfill\hbox{${\vcenter{\vbox{
    \hrule height 0.4pt\hbox{\vrule width 0.4pt height 6pt
    \kern5pt\vrule width 0.4pt}\hrule height 0.4pt}}}$}}

\def\tcr{\textcolor{red}}
\def\tcb{\textcolor{blue}}

\definecolor{myred}{rgb}{0.3,0.0,0.7}
\definecolor{dkg}{rgb}{0.1,0.7,0.2}
\definecolor{dkb}{rgb}{0.0,0.2,0.8}

 \def\hL{\widehat{L}}
 \def\hM{\widehat{M}}

 \def\hS{\widehat{S}}

 \def\hX{\widehat{X}}

\def\Ac{{\cal A}}

\def\Cc{{\cal C}}

\def\Pbb{{\mathbb P}}

\def\Rbb{{\mathbb R}}

\newcommand{\bprfof}{\begin{proof_of}}
\newcommand{\eprfof}{\end{proof_of}}
\newcommand{\bprf}{\begin{myproof}}
\newcommand{\eprf}{\end{myproof}}
\newcommand{\bp}{\begin{psfrags}}
\newcommand{\ep}{\end{psfrags}}
\newcommand{\bl}{\begin{lemma}}
\newcommand{\el}{\end{lemma}}
\newcommand{\bt}{\begin{theorem}}
\newcommand{\et}{\end{theorem}}
\newcommand{\bc}{\begin{center}}
\newcommand{\ec}{\end{center}}
\newcommand{\bi}{\begin{itemize}}
\newcommand{\ei}{\end{itemize}}
\newcommand{\ben}{\begin{enumerate}}
\newcommand{\een}{\end{enumerate}}
\newcommand{\bd}{\begin{definition}}
\newcommand{\ed}{\end{definition}}
\def\beq{\begin{equation}}
\def\eeq{\end{equation}\noindent}
\def\beqn{\begin{eqnarray}}
\def\eeqn{\end{eqnarray} \noindent}
\def\beqnn{  \begin{eqnarray*}}
\def\eeqnn{\end{eqnarray*}  \noindent}
\def\bcase{  \begin{numcases}}
\def\ecase{\end{numcases}   \noindent}
\def\bsbcase{  \begin{subnumcases}}
\def\esbcase{\end{subnumcases}   \noindent}

\newenvironment{myproof}{\noindent{\bf Proof:} \hspace*{1em}}{
    \hspace*{\fill} $\Box$ }
\newenvironment{proof_of}[1]{\noindent {\bf Proof of #1: }}{\hspace*{\fill} $\Box$ }

\newcommand{\matplottc}[1]{               
        \unitlength .45truein
        \begin{center}
        \includegraphics{#1.ps}
        \end{picture}
        \end{center}
}

\def\psfancypar#1#2{\begingroup\def\par{\endgraf\endgroup\lineskiplimit=0pt}
               \setbox2=\hbox{\large\sc #2}
               \newdimen\tmpht \tmpht \ht2 \advance\tmpht by \baselineskip
               \font\hhuge=Times-Bold at \tmpht
               \setbox1=\hbox{{\hhuge #1}}
               \count7=\tmpht \count8=\ht1
               \divide\count8 by 1000 \divide\count7 by \count8
               \tmpht=.001\tmpht\multiply\tmpht by \count7
               \font\hhuge=Times-Bold at \tmpht
               \setbox1=\hbox{{\hhuge #1}}
               \noindent
                \hangindent1.05\wd1
               \hangafter=-2 {\hskip-\hangindent
               \lower1\ht1\hbox{\raise1.0\ht2\copy1}%
                \kern-0\wd1}\copy2\lineskiplimit=-1000pt}

\def\Kout{\setbox1=\hbox{\Huge\bf K}\hbox to
1.05\wd1{\hspace{.05\wd1}
}}
\def\Sout{\setbox1=\hbox{\Huge\bf S}\hbox to 1.05\wd1{\hspace{.05\wd1}
}}

\title{A Clustering Approach to Learn Sparsely-Used Overcomplete
  Dictionaries}
\date{}

\author{Alekh Agarwal, Animashree Anandkumar, and Praneeth
  Netrapalli\,\footnote{A. Agarwal is with Microsoft Research, New
    York. Email: alekha@microsoft.com. A. Anandkumar is with the
    Center for Pervasive Communications and Computing, Electrical
    Engineering and Computer Science Dept., University of California,
    Irvine, USA 92697. Email: a.anandkumar@uci.edu. P. Netrapalli is
    with Dept. of ECE, The University of Texas at Austin. Email:
    praneethn@utexas.edu.  The work was initiated during the visits of
    A. Anandkumar and P. Netrapalli to Microsoft Research New England
    in Summer 2013. A preliminary version of this manuscript
    containing a subset of these results appears in the proceedings of
    COLT 2014.}}

\begin{document}
\maketitle

\begin{abstract}
We consider the problem of learning overcomplete dictionaries in the
context of sparse coding, where each sample selects a sparse subset of
dictionary elements. Our main result is a strategy to approximately
recover the unknown dictionary using an efficient algorithm. Our
algorithm is a clustering-style procedure, where each cluster is used
to estimate a dictionary element. The resulting solution can often be
further cleaned up to obtain a high accuracy estimate, and we provide
one simple scenario where $\ell_1$-regularized regression can be used
for such a second stage. 
\end{abstract}



\section{Introduction}

The dictionary learning problem is as follows: given observations $Y$,
the task is to factorize it as \beq\label{eqn:dict} Y= AX,\quad Y\in
\Rbb^{d \times n}, \,\, A\in \Rbb^{d \times r},\,\,X\in \Rbb^{r \times
  n},\eeq where $X$ is referred to as the {\em coefficient} matrix and
the columns of $A$ are referred to as the {\em dictionary} elements.
There are indeed infinite factorizations for \eqref{eqn:dict} unless
further constraints are imposed. A natural assumption is that the
coefficient matrix $X$ is sparse, and in fact, that each sample $y_i$
selects a sparse subset of dictionary elements from $A$. This instance
of dictionary learning is popularly known as the {\em sparse coding}
problem~\cite{olshausen1997sparse,lee2006efficient}. It has been
argued that sparse coding can provide a succinct representation of the
observed data, given only unlabeled
samples~\cite{lee2006efficient}. Through this lens of unsupervised
learning, dictionary learning has received an increased attention from
the machine learning community in the last few years; see
Section~\ref{sec:relatedwork} for a brief survey.

Although the above problem has been extensively studied, most of the
methods are heuristic and lack guarantees.  Spielman
et. al~\cite{spielman2012exact} provide exact recovery results for
this problem, when the coefficient matrix has Bernoulli-Gaussian
entries and the dictionary matrix $A\in \Rbb^{r\times d}$ has full
column rank. This condition entails that the dictionary is {\em
  undercomplete}, i.e., the observed dimensionality needs to be
greater than the number of dictionary elements $(r \leq d)$. However,
for most practical settings, it has been argued that {\em
  overcomplete} representations, where $r \gg d$, are far more
relevant, and can provide greater flexibility in modeling as well as
greater robustness to
noise~\cite{lewicki2000learning,bengio2012unsupervised}. Moreover, in
the context of blind source separation (BSS) of audio, image or video
signals, the dictionary learning problem is typically overcomplete,
since there are more sources than observations~\cite{elad2010sparse}.
In this work, we provide guaranteed methods for learning overcomplete
dictionaries.

\subsection{Summary of Results}

In this paper we present a novel algorithm for the estimation of
overcomplete dictionaries. The algorithm can be seen as a
\emph{clustering style method} followed by a singular value
decomposition (SVD) within each cluster resulting in an estimate for
each dictionary element. The clusters are formed based on the
magnitudes of the correlation between pairs of samples. Under our
probabilistic model of generating data as well as assumptions on the
coefficients and dictionaries, it can be guaranteed that such a
procedure approximately recovers the unknown overcomplete
dictionary. Under further conditions, it is often possible to start
with this approximate solution and perform additional post-processing
on it to obtain arbitrarily good estimates of the dictionary. We
present one such set of conditions under which \emph{sparse
  regression} can be used for this post-processing. More advanced
post-processing methods have been developed in subsequent
works~\cite{AgarwalAJNT13,Arora2013}.

We consider a random coefficient matrix, where each column of $X$ has
$s$ non-zero entries which are randomly chosen, i.e., each sample
$y_i$ selects $s$ dictionary elements uniformly at random. We
additionally assume that the dictionary elements are pairwise {\em
  incoherent} and that the dictionary matrix satisfies a certain bound
on the spectral norm. Under these conditions, we establish that our
algorithm estimates the dictionary elements with bounded (constant)
error when the number of samples scales as $n=\order(r(\log r+ \log
d))$, and when the sparsity $s = \order(d^{1/4},r^{1/4})$. To the best
of our knowledge, this is the first result of its kind which analyzes
the global recovery properties of a computationally efficient
procedure in the setup of overcomplete dictionary learning.

In the special case when the coefficients are $\{-1, 0, 1\}$-valued
with zero mean, the resulting solution from the first step can be
further plugged into any sparse regression algorithm for estimating
the coefficients given this dictionary estimate. Under a more
stringent sparsity constraint: $s = \order(d^{1/5},r^{1/6})$, it can
be shown that this second step will recover the coefficients
\emph{exactly} even from this approximate dictionary, which then also
leads to an exact recovery of the dictionary by solving the linear
system. Hence, we provide a simple method for exactly recovering the
unknown dictionary in this special case. A natural generalization
of this procedure to general weights is analyzed using alternating minimization procedure in a subsequent
work~\cite{AgarwalAJNT13}.

We outline our method as well as our analysis techniques in
Section~\ref{sec:overview}. This is the first work to provide a
tractable method for {\em guaranteed recovery} of overcomplete
dictionaries, and we discuss the previous results below. Finally,
concurrently with our work, an approximate recovery result with a
similar procedure was recently announced by Arora et
al.~\cite{Arora2013}. A detailed discussion comparing our and their
results is presented in Section~\ref{sec:relatedwork}.

\subsection{Related Works}
\label{sec:relatedwork}

This work overlaps with and relates to prior works in many different
communities and we discuss them below in turn.

\paragraph{Dictionary Learning: }
Hillar and Sommer~\cite{hillar2011ramsey} consider conditions for
identifiability of sparse coding and establish that when the
dictionary succeeds in reconstructing a certain set of sparse vectors,
there exists a unique sparse coding, up to permutation and
scaling. However, the number of samples required to establish
identifiability is exponential in $r$ for the general case. In
contrast, we show that efficient recovery is possible using
$\order(r(\log r + \log d))$ samples, albeit under additional
conditions such as {\em incoherence} among the dictionary elements.

Spielman et. al~\cite{spielman2012exact} provide exact recovery
results for a $\ell_1$ based method in the {\em undercomplete}
setting, where $r \leq d$. In contrast, we allow for the overcomplete
setting where $r> d$. There exist a plethora of heuristics for
dictionary learning, which work well in practice in many contexts, but
lack theoretical guarantees. For instance, Lee et. al. propose an
iterative $\ell_1$ and $\ell_2$ optimization
procedures~\cite{lee2006efficient}. This is similar to the the method
of optimal directions (MOD) proposed in~\cite{engan1999method}.
Another popular method is the so-called K-SVD, which iterates between
estimation of $X$ and given an estimate of $X$, updates the dictionary
estimate using a spectral procedure on the residual. Other works
consider more sophisticated methods from an optimization viewpoint
while still alternating between dictionary and coefficient
updates~\cite{jenatton2010proximal,GengWaWr2011}. Geng et
al.~\cite{GengWaWr2011} and Jenatton et al.~\cite{jenatton2012local}
study the local optimality properties of an alternating minimization
procedure. In contrast, our work focuses on global properties of a
more combinatorial procedure than several of the above works which are
more optimization flavored. The upshot is that our procedure, while
still being computationally quite efficient, is able to guarantee
global bounds on the quality of the solution obtained.



Recent works~\cite{vainsencher2011sample,
  mehta2013sparsity,maurer2012sparse,TRS:dict2013} provide
generalization bounds and algorithmic stability for predictive sparse
coding, where the goal of the learned sparse representation is to
obtain good performance on some predictive task. This differs from our
framework since we do not consider predictive tasks here, but the
accuracy in recovering the underlying dictionary elements.

Finally, our results are closely related to the very recent work of
Arora et al.~\cite{Arora2013}, carried out independently and
concurrently with our work.  There are however some important
distinctions: we require only $\order(r)$ samples in our analysis,
while Arora et al.~\cite{Arora2013} require $\order(r^2)$ samples in
their result. At the same time, their analysis yields milder
conditions on the sparsity level $s$ in terms of its dependence on $r$
and $d$. Following this work, Arora et al.~\cite{Arora2013} and
Agarwal et al.~\cite{AgarwalAJNT13} also developed a post-processing
techniques which can be thought of as a more advanced variant of the
simpler sparse-regression step that we analyze. These subsequent works
view the methods developed here as initialization procedures to
alternating optimization schemes.


\paragraph{Blind Source Separation/ICA/Topic Models: }
The problem of dictionary learning is applicable to blind source
separation (BSS), where the rows of $X$ are signals from the sources
and $A$ represents the linear mixing matrix. The term {\em blind}
implies that the dictionary matrix $A$ is unknown and needs to be
jointly estimated with the coefficient matrix $X$, given samples
$Y$. This problem has been extensively studied and the most popular
setting is the independent component analysis (ICA), where the sources
are assumed to be independent. In contrast, for the sparse component
analysis problem, no assumptions are made on the statistics of the
sources.  Many works provide guarantees for ICA in the undercomplete
setting, where there are fewer sources than
observations~\cite{hyvarinen2000independent,arora2012provable,AnandkumarEtal:tensor12}
and some works provide guarantees in the overcomplete
setting~\cite{de2007fourth,FourierPCA}. However, the techniques are
very different since they rely on the independence among the sources.
The problem of learning topic models can be cast as a similar
factorization problem, where $A$ now corresponds to the topic-word
matrix and $X$ corresponds to the proportions of topics in various
documents. There are various recent works providing guaranteed methods
for learning topic models,
e.g~\cite{AnandkumarEtal:ldaNIPS12,AroraICML,AnandkumarEtal:DAG12,AnandkumarEtal:overcomplete13}. However,
these works make different assumptions on either $A$ or $X$ or both to
guarantee recovery. For instance, the work~\cite{AroraICML} assumes
that the topic-word matrix $A$ has rows such that for each column,
only the entry corresponding to that column is non-zero. The
work~\cite{AnandkumarEtal:DAG12} assumes expansion conditions on $A$
and provides recovery through $\ell_1$-based optimization. We note
that the techniques of~\cite{AnandkumarEtal:DAG12} are related to
those employed by Spielman et. al~\cite{spielman2012exact} for
dictionary learning, but make different assumptions. All these works
only deal with the undercomplete setting. The recent
work~\cite{AnandkumarEtal:overcomplete13} considers topic models in
the overcomplete setting, and provides guarantees when $A$ satisfies
certain higher order expansion conditions. The techniques are very
different from the ones employed here since they involve higher order
moments and tensor forms.

\paragraph{Connection to Learning Overlapping Communities: }
Our initial step for estimating the dictionary elements involves
finding large cliques in the sample correlation graph, where the nodes
are the samples and the edges represent sufficiently large
correlations among the endpoints.  The clique finding problem is a
special instance of the overlapping community detection problem, which
has been studied in various contexts,
e.g.~\cite{AnandkumarEtal:community12COLT,BBBCT12,AGSS12,ChenSanghaviXu,McSherry01}.
However, the correlation graph here has different kinds of constraints
than the ones studied before as follows. In our setting involving
noise-free dictionary learning, each community corresponds to a clique
and there are no edges across two different communities. In contrast,
many works on community detection are concerned about handling {\em
  noise} efficiently, where each community is not a full clique, and
there are edges across different communities. Here, we need to learn
overlapping communities, while many community detection methods limit
to learning non-overlapping ones. In our setting, we argue that the
overlap across different communities is small under a random
coefficient matrix, and thus, we can find the communities efficiently
through simple random sampling and neighborhood testing procedures.


\subsection{Overview of Techniques}
\label{sec:overview}

As stated earlier, our main algorithm consists of a clustering
procedure which yields an approximate estimate of the dictionary. This
estimate can be subsequently post-processed for exact recovery of the
dictionary under certain further conditions. Below we give the outline
and the main intuition underlying these procedures and their
analysis.

\paragraph{Dictionary estimation via clustering: }
This step first involves construction of the sample correlation graph
$G_{\corr(\rho)}$, where the nodes are samples $\{y_1, y_2, \ldots
y_n\}$ and an edge $(y_i, y_j)\in G_{\corr(\rho)}$ implies that
$|\inner{y_i, y_j}| > \rho$, for some $\rho>0$. We then employ a {\em
  clustering} procedure on the graph to obtain a subset of samples,
which are then employed to estimate each dictionary element. Roughly,
we search for large cliques in the correlation graph and obtain a
spectral estimate of each dictionary element using samples from such
sets.

\begin{figure}
\subfloat[a][Coefficient bipartite graph $B$ mapping dictionary
  elements $a_1, a_2, \ldots a_r$ to samples $y_1, \ldots y_n$: $y_i =
  \sum_{j \in [r]} x_{ji} a_j$. See \eqref{eqn:cc-intro} for
  definition of $\Cc_i$.]
         {\begin{minipage}{3in}\label{fig:bipartite}\centering
             \bp\psfrag{k}[c]{$r$}\psfrag{d}[c]{$n$}
             \psfrag{h1}[c]{$a_1$}\psfrag{h2}[c]{$a_2$}
             \psfrag{h3}[c]{$a_r$}\psfrag{x1}[c]{$y_1$}
             \psfrag{x2}[c]{$y_2$}\psfrag{x4}[c]{ } \psfrag{x5}[c]{
             }\psfrag{x6}[c]{$y_n$} \psfrag{A}[c]{$X$}
             \psfrag{C1}[c]{\tcr{$\Cc_1$}}
             \psfrag{Cr}[l]{\tcr{$\Cc_r$}}
             \psfrag{C12}[l]{\tcb{$\Cc_{12}$}}
             \centering{\includegraphics[width=2.5in,height=1.5in]{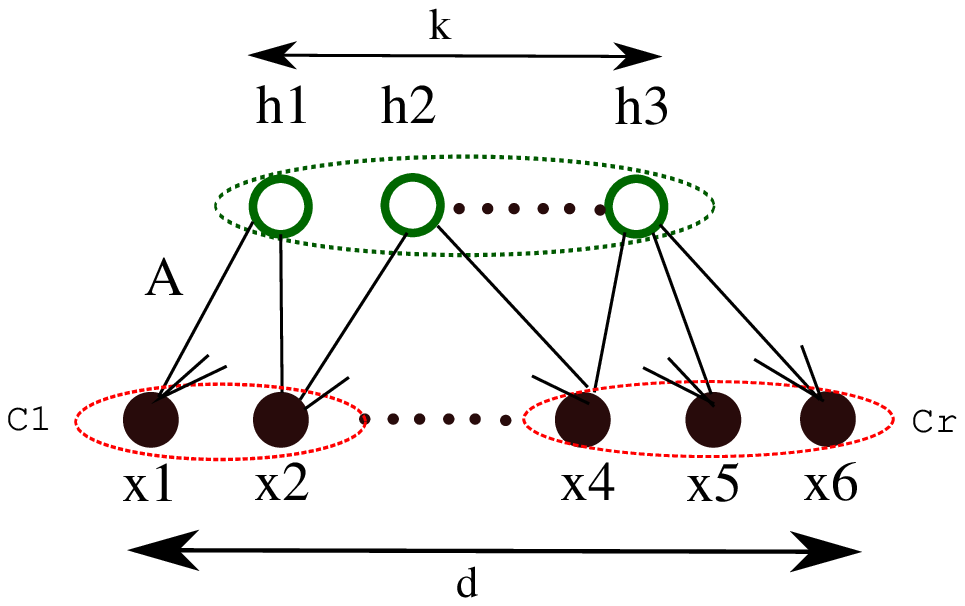}}
             \ep\end{minipage}}
         \hfil \subfloat[b][Sample correlation graph $G_{\corr}$ with
           nodes $\{y_k\}$ and edge $(y_i, y_j)$ s.t. $|\inner{y_i,
             y_j}| > \rho$.  $\Cc_i$, defined in \eqref{eqn:cc-intro},
           is a clique in the correlation graph. See
           Lemma~\ref{lemma:corrgraph}.]
         {\begin{minipage}{3in}\label{fig:clustergraph}\centering
             \bp \psfrag{Ci}[l]{$\Cc_i$} \psfrag{Ci}[l]{$\Cc_i$}
             \psfrag{Cj}[c]{$\Cc_j$} \psfrag{Cj}[l]{$\Cc_j$}
             \centering{
               \includegraphics[width=2.5in,height=1.5in]{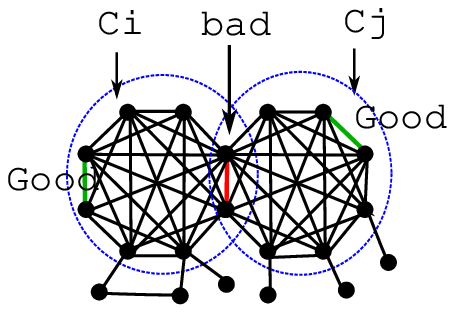}}
             \ep\end{minipage}}
         \caption{Coefficient bipartite graph and the sample correlation graph.}
  \label{fig:bipartite-corr} \end{figure}

\paragraph{Key intuitions for the clustering procedure:}
The core intuitions can be described in terms of the relationships
between the two graphs, \viz the coefficient bipartite graph
$B_{\coeff}$ and the sample correlation graph $G_{\corr}$, shown in
Figures~\ref{fig:bipartite} and~\ref{fig:clustergraph}. As described
earlier, the correlation graph $G_{\corr}$ consists of edges between
well correlated samples.  The coefficient bipartite graph $B_{\coeff}$
consists of dictionary elements $\{a_i\}$ on one side and the samples
$\{y_i\}$ on the other, and the bipartite graph $B_{\coeff}$ encodes
the sparsity pattern of the coefficient matrix $X$. In other words, it
maps the dictionary elements $\{a_i\}$ to samples $\{y_i\}$ on which
they are supported on and $\nbd_B(y_i)$ denotes the neighborhood of
$y_i$ in the graph $B_{\coeff}$.

Now given this bipartite graph $B_{\coeff}$, for each dictionary
element $a_i$, consider a set of samples\footnote{Note that such a set
  need not be unique.} which (pairwise) have only one dictionary
element $a_i$ in common, and denote such a set by $\Cc_i$ i.e.
\beq \label{eqn:cc-intro} \Cc_i := \{ y_k, k\in S: \nbd_B(y_k)\cap
\nbd_B(y_l)=a_i, \,\, \forall\,k,l\in S\}.\eeq For a random
coefficient matrix (resulting in a random bipartite graph), we argue
that there exists (large) sets $\Cc_i $, for each $i\in [r]$, which
consists of a large fraction of $\nbd_B(a_i)$, and no two elements
$a_i$ and $a_j$ have a large fraction of samples in common. In other
words, for random coefficient matrices, we see a diversity in the
dictionary elements among the samples, and this can be viewed as an
{\em expansion} property from the dictionary elements to the set of
samples. We exploit this property to establish success for our method.

Our subsequent analysis is broadly divided into two parts, \viz
establishing that (large) sets $\{\Cc_i\}$ can be found efficiently,
and that the dictionary elements can be estimated accurately once such
sets $\{\Cc_i\}$ are found.  We establish that the sets $\{\Cc_i\}$
are cliques in the correlation graph when the dictionary elements are
incoherent, as shown in Figure~\ref{fig:clustergraph}. Combined with
the previous argument that the different sets $\Cc_i$'s have only a
small amount of overlap for random coefficient matrices, we argue that
these sets can be found efficiently through simple random sampling and
neighborhood testing on the correlation graph.  Once a large enough
set $\Cc_i$ is found, we argue that under incoherence, the dictionary
element $a_i$ can be estimated accurately through SVD over the samples
in $\Cc_i$.

\paragraph{Sparse regression for post-processing:}
This is a relatively straightforward procedure. Once an initial
estimate of the dictionary matrix is obtained, we estimate the
coefficient matrix $X$ through any sparse regression procedure (such
as Lasso) and then perform thresholding on the recovered coefficients.
Now, we re-estimate the dictionary, given this coefficient matrix, by
solving another linear system. This provides us with a final estimate
of both the dictionary as well as the coefficient matrix.

Since we only have a noisy estimate of the dictionary, our analysis
here is slightly different from the usual analysis for a sparse linear
system. The noise in our system is dependent on the approximate
dictionary employed, which differs from the typical statistical
setting, where noise is assumed to be independent.  We exploit the
known guarantees available for Lasso under deterministic
noise~\cite{Candes2008} for our setting. Combining Lasso with a simple
thresholding procedure, we guarantee exact recovery of the coefficient
matrix, albeit under a more stringent condition on the sparsity and
the coefficient values (namely zero mean and $\{-1, 0, 1\}$-valued
). The dictionary is then re-estimated by solving another linear
system, which is of course correct owing to the exact estimation of
the coefficient matrix.

\section{Method and Guarantees}

\paragraph{Notation: }
Let $[n]:=\{1,2, \ldots, n\}$ and for a vector $w$, let $\supp(w)$
denote the support of $w$, i.e. the set of indices where $w$ is
non-zero. Let $\|w\|$ denote the $\ell_2$ norm of vector $w$, and
similarly for a matrix $W$, $\|W\|$ denotes its spectral norm. Let
$A=[a_1|a_2|\ldots |a_r]$, where $a_i$ denotes the $i^{\tha}$ column,
and similarly for $Y=[y_1| y_2|\ldots y_n]$ and
$X=[x_1|\ldots|x_n]$. For a graph $G=(V,E)$, let $\nbd_G(i)$ denote
set of neighbors for node $i$ in $G$.

\subsection{Clustering procedure and its analysis}\label{sec:method}

We start with presenting the main algorithm of our work and bound the
recovery error under certain assumptions. 

\subsubsection{Algorithm}

Our main algorithm is presented in
Algorithm~\ref{algo:init-dict}. Given samples $Y$, we first construct
the correlation graph $G_{\corr(\rho)}$, where the nodes are samples
$\{y_1, y_2, \ldots y_n\}$ and an edge $(y_i, y_j)\in G_{\corr(\rho)}$
implies that $|\inner{y_i, y_j}| > \rho$, for some threshold
$\rho>0$. We then determine a good subset of samples via a {\em
  clustering} procedure on the graph as follows: we first randomly
sample an edge $(y_{i^*}, y_{j^*})\in G_{\corr(\rho)}$ and then
consider the intersection of their neighborhoods, denoted by $\hS$. We
then employ UniqueIntersection routine in
Procedure~\ref{procedure:uniqueint} to determine if $\hS$ is a ``good
set'' for estimating a dictionary element, and this is done by
ensuring that the set $\hS$ has sufficient number of mutual
neighbors\footnote{For convenience to avoid dependency issues, in
  Procedure~\ref{procedure:uniqueint}, we partition $\hS$ into sets
  consisting of node pairs and determine if there are sufficient
  number of node pairs which are neighbors.}  in the correlation
graph. Once $\hS$ is determined to be a good set, we then proceed by
estimating the matrix $\hM$ using samples in $\hS$ and output its top
singular vector as the estimate of a dictionary element. The method is
repeated over all edges in the correlation graph to ensure that all
the dictionary elements get estimated with high probability.

\begin{algorithm}[t]
\caption{DictionaryLearn$(Y, \epsilon_{\dict}, \rho)$: Clustering approach for estimating dictionary elements.}\label{algo:init-dict}
\begin{algorithmic}
\renewcommand{\algorithmicrequire}{\textbf{Input: }}
\renewcommand{\algorithmicensure}{\textbf{Output: }}
\REQUIRE Samples $Y=[y_1| \ldots| y_n]$. Correlation threshold
$\rho$. Desired separation parameter $\epsilon$ between recovered
dictionary elements.
\ENSURE  Initial Dictionary Estimate $\Ainit$.
\STATE Construct correlation graph $G_{\corr(\rho)}$
s.t. $(y_i,y_j)\in G_{\corr(\rho)}$ when $|\inner{y_i,y_j}|>\rho$.
\STATE  Set $\Ainit\leftarrow \emptyset$.
\FOR{each edge $(y_{i^*}, y_{j^*}) \in G_{\corr(\rho)}$}
\STATE $\hS\leftarrow \nbd_{G_{\corr(\rho)}}(y_{i^*})\cap \nbd_{G_{\corr(\rho)}}(y_{j^*})$.
\IF{UniqueIntersection$(\hS, G_{\corr(\rho)})$}
\STATE $\hL \leftarrow \sum_{y\in \hS} y y^\top$ and $\ainit\leftarrow
u_1$, where $u_1$ is top singular vector of $\hL$.
\IF{$\min_{b\in \Ainit}\|\ainit-b\| > 2\epsilon_{\dict}$}
\STATE $\Ainit\leftarrow \Ainit\cup \ainit$
\ENDIF
\ENDIF
\ENDFOR
\STATE Return $\Ainit$
\end{algorithmic}
\end{algorithm}

\floatname{algorithm}{Procedure}
\setcounter{algorithm}{0}
\begin{algorithm}[t]
\caption{UniqueIntersection$(S, G)$: Determine if samples in $S$ have a unique intersection.}\label{procedure:uniqueint}
\begin{algorithmic}
\renewcommand{\algorithmicrequire}{\textbf{Input: }}
\renewcommand{\algorithmicensure}{\textbf{Output: }}
\REQUIRE Set $S$ with $2\subsamp$ vectors $y_1, \ldots y_{2\subsamp}$ and graph $G$
with $y_1, \ldots, y_{2\subsamp}$ as nodes.
\ENSURE  Indicator variable UNIQUE\_INT
\STATE Partition $S$ into sets $S_1, \ldots, S_{\subsamp}$ such that each $|S_t|=2$.
\IF{$\left|\{t \middle\vert S_t \in G\}\right| > \frac{61\subsamp}{64}$}
\STATE  UNIQUE\_INT $\leftarrow 1$
\ELSE
\STATE UNIQUE\_INT $\leftarrow 0$
\ENDIF
\STATE Return UNIQUE\_INT
\end{algorithmic}
\end{algorithm}

\subsubsection{Assumptions and Main Result}

\paragraph{Assumptions: }
We now provide guarantees for the proposed method under the following
assumptions on $A$ and $X$.

\ben\item[$(A1)$] {\bf Unit-norm Dictionary Elements: }All the
elements are normalized: $\|a_i\| =1$, for $i \in [r]$.

 \item[$(A2)$] {\bf Incoherent Dictionary Elements: }We assume
   pairwise incoherence condition on the dictionary elements, for some
   constant $\mu_0>0$,

   \beq \label{eqn:incoherence} |\inner{a_i, a_j}| <
   \frac{\mu_0}{\sqrt{d}}.\eeq

\item[$(A3)$] {\bf Spectral Condition on Dictionary Elements: }The
  dictionary matrix has bounded spectral norm, for some constant
  $\mu_1>0$, 
  \beq \label{eqn:spectral} \|A \| < \mu_1\sqrt{\frac{r}{d}}.\eeq

\item[$(A4)$] {\bf Entries in Coefficient Matrix: }We assume that the
  non-zero entries of $X$ are drawn from a zero-mean distribution
  supported on $\left[-M, -m\right]\cup\left[m,M\right]$ for some
  fixed constants $m$ and $M$.

\item[$(A5)$] {\bf Sparse Coefficient Matrix: } The columns of
  coefficient matrix have bounded number of non-zero entries $s$ which
  are selected randomly, i.e.  
  \beq \label{eqn:sparsity} |\supp(x_i)| =s, \quad \forall \, i \in
     [n].\eeq

    We require $s$ to be
    
    \beq \label{eqn:sparsitylevel} s< c
  \min\left(\sqrt{\frac{m^2\sqrt{d}}{2 M^2\mu_0}},
  \sqrt[3]{\frac{r}{1536}}\right), \eeq 
  for some small enough constant $c$.

\item[$(A6)$] {\bf Sample Complexity: }Given a parameter $\alpha\in
  (0,1/20)$ (which is related to the error in recovery of dictionary,
  see Theorem~\ref{thm:main-init}), and a universal constant $c>0$,
  choose $\delta>0$ and the number of samples $n$ such that

  \[
  n := n(d, r, s, \delta, \alpha)= \frac{cr}{\alpha^2 s}\log
  \frac{d}{\delta},  \quad n^2 \delta<1.
  \]

\item[$(A7)$] {\bf Choice of Threshold for Correlation Graph: }The
  correlation graph $G_{\corr(\rho)}$ is constructed using threshold
  $\rho$ such that

  \beq \label{eqn:rho} \rho= \frac{m^2}{2} -
  \frac{s^2 M^2 \mu_0}{\sqrt{d}}>0. \eeq

\item[$(A8)$] {\bf Choice of Separation Parameter $\epsilon_{\dict}$
  between Estimated Dictionary Elements:} This is the desired accuracy
  of the estimated dictionary elements to the true dictionary elements
  using just the initialization step. It can be chosen to be:

  \beq
  \frac{32sM^2}{m^2}\left(\frac{\mu_1}{\sqrt{ds}}+ \frac{\mu_1^2}{d} +
  \frac{s^3}{r} + \alpha^2 + \frac{\alpha}{\sqrt{s}}\right) <
  \epsilon^2_{\dict} < \frac{1}{4}.
  \eeq
  \een

The assumption $(A1)$ on normalization is without loss of generality
since we can always rescale the dictionary elements and the
corresponding coefficients and obtain the same observations. The
assumption $(A2)$ on incoherence is crucial to our analysis.  In
particular, incoherence also leads to a bound on the RIP constant; see
Lemma~\ref{lem:RIP-A} in Appendix~\ref{app:aux}.  The assumption
$(A3)$ provides a bound on the spectral norm of $A$.

The assumption $(A4)$ assumes that the non-zero entries of $X$ are
drawn from a zero-mean distribution with natural upper and lower
bounds on the coefficients. Note that a similar assumption is made in
the work of Arora et.al~\cite{Arora2013}. 

The assumption $(A5)$ on sparsity in the coefficient matrix is crucial
for identifiability of dictionary learning problem. We require for the
sparsity to be not too large for recovery.

The assumption $(A6)$ provides a bound on sample complexity. We
subsequently establish that in order to have decaying error for
recovery of dictionary elements, we require $n = \omega(r)$ samples
for recovery. Thus, we obtain a nearly linear sample complexity for
our method.


Assumption $(A7)$ specifies the threshold for the construction of the
correlation graph. Intuitively, we require a threshold such that we
can distinguish pairs of samples which share a dictionary element from
those which do not.


\paragraph{Main Result:} We now present our main result which bounds
the error in the estimates of Algorithm~\ref{algo:init-dict}. 

\begin{theorem}[Approximate recovery of dictionary]
\label{thm:main-init}
Suppose the output of Algorithm~\ref{algo:init-dict} is $\Ainit$. Then
with probability greater than $1-2n^2\delta$, there exists a
permutation matrix $P$ such that:
\begin{align}\label{eqn:approx}
\errA^2:= \min_{i\in [r]} \min_{z\in \{-1,+1\}}\twonorm{za_i -
  (P\Ainit)_i}^2 < \frac{32sM^2}{m^2}\left(\frac{\mu_1}{\sqrt{ds}}+
\frac{\mu_1^2}{d} + \frac{s^3}{r} + \alpha^2 +
\frac{\alpha}{\sqrt{s}}\right).
\end{align}
\end{theorem}

\paragraph{Remark: } 
Note that we have a sign ambiguity in recovery of the dictionary
elements, since we can exchange the signs of the dictionary elements
and the coefficients to obtain the same observations.  The assumption
on sparsity in $(A4)$ implies that the first two terms
in~\eqref{eqn:approx} decay. For the third term in~\eqref{eqn:approx}
to decay, we require $s=o(r^{1/4})$ instead of $s=\order(r^{1/3})$ as
in $(A4)$. Moreover, we require that $\alpha^2 s = o(1)$. Since the
sample complexity in $(A7)$ scales as $n =
\Omega\left(\frac{r}{\alpha^2 s}\right)$, we require $n = \omega(r)$
samples for recovery of dictionary with decaying error. Thus, we
obtain a near linear sample complexity for our method. We observe that
the error in our estimation depends inversely on dimension-related
quantities such as $d$ and $r$ and not on the number of samples
$n$. This is because the errors in our estimates arise from errors in
SVD step, specifically from the discrepancy between the SVD vector and
the dictionary element responsible for a cluster. Even the population
SVD will suffer from an approximation error here, which is responsible
for our error bound, but the probability in the error bound improves
with the number of samples as we get closer and closer to the
population SVD estimate.

\subsection{Post-processing for binary coefficients} \label{sec:post-process} 

We now present the post-processing step which will be analyzed under a
more stringent condition on the coefficients.

\subsubsection{Algorithm}

\floatname{algorithm}{Algorithm}
\setcounter{algorithm}{1}
\begin{algorithm}[t]
\caption{RecoverCoeff$(Y, \Ainit, \epsilon_{\coeff})$: Exact Recovery through lasso}
\label{algo:lasso}
\begin{algorithmic}[1]
\INPUT Samples $Y$, approximate dictionary $\Ainit$ and accuracy
parameter $\epsilon_{\coeff}$. $\sign(X)$ returns a matrix with signs of the entries of $X$.
\FOR{samples $i = 1,2,\ldots, n$}
\STATE Estimate
\begin{equation}
  \widehat{x}_i = \arg\min_{x \in \R^r} \norm{x}_1, \quad \mbox{subject
    to} \quad \norm{y_i - \Ainit x}_2 \leq \epsilon_{\coeff}.
  \label{eqn:lasso}
\end{equation}
\STATE Threshold: $\widehat{X}\leftarrow \sign(\hX)$.
\ENDFOR
\STATE Estimate $\widehat{A} =
Y\widehat{X}^T(\widehat{X}\widehat{X}^T)^{-1}$
\STATE Normalize: $\widehat{a}_i =
\frac{\widehat{a}_i}{\norm{\widehat{a}_i}_2} $
\OUTPUT $\widehat{A}$
\end{algorithmic}
\end{algorithm}

Once we obtain an estimate of the dictionary elements, we proceed to
estimate the coefficient matrix. The main observation at this step is
that the coefficient vector $x_i$ for each sample $y_i$ is a
$s$-sparse vector in $r$-dimensions. Hence, recovering the
coefficients would be a standard sparse linear problem if we knew the
dictionary $A$ exactly. Our analysis will show that even an
approximately correct dictionary $\Ainit$ from
Algorithm~\ref{algo:init-dict} suffices to provide guarantees for this
recovery. Once the coefficients are estimated, the dictionary can be
re-estimated by solving another linear system. The procedure is
formally described in Algorithm~\ref{algo:lasso}. We do not prescribe
any particular choice of computational procedure to solve the
optimization problem~\eqref{eqn:lasso}, but there are many algorithms
available in standard literature. As a concrete example, the GraDeS
algorithm of Garg and Khandekar~\cite{GargK09} or OMP of Tropp and
Gilbert~\cite{Tropp2007}works in our setting.

\subsubsection{Exact recovery for bernoulli coefficients}

Our second result is that under stronger conditions than before, it is
possible to exactly recover the unknown dictionary $A$ with high
probability. This result will be obtained by initializing
Algorithm~\ref{algo:lasso} with the output of
Algorithm~\ref{algo:init-dict}. We start with the additional
assumptions, putting restrictions on the allowed sparsity level $s$ as
a function of $r$ and $d$.

\begingroup
\def\theassumption{B1}
\begin{assumption}[Conditions for exact recovery]
  The non-zero coefficients in coefficient matrix $X$ are zero-mean
  Bernoulli$\{-1,1\}$. This corresponds to setting $M = m = 1$ in
  Assumption (A4).

  The sparsity level $s$, and the number of dictionary elements $r$
  and the observed dimension $d$ satisfy
  \begin{align*}
  32s\left( \frac{\mu_1}{\sqrt{ds}} + \frac{\mu_1^2}{d}\right) \leq
  \frac{1}{1200s^2}, \quad \mbox{and}\quad
  \frac{32s^4}{r} \leq \frac{1}{1200s^2}.
  \end{align*}The constant
  $\alpha$ in Theorem~\ref{thm:main-init} satisfies
\begin{align*}
    32s\left( \alpha^2 + \frac{\alpha}{\sqrt{s}} \right) &\leq
    \frac{1}{1200s^2}.
  \end{align*}
The number of samples $n$, in addition to assumption $(A6)$, satisfies

  \[ n \geq \frac{4r}{c_0}\log \frac{d}{\delta},\]
  where $c_0$ is a universal constant.

  The accuracy parameter $\epsilon_{\coeff}$ in
  Algorithm~\ref{algo:lasso} is chosen as $\epsilon_{\coeff}= s
  \errA$, where $\errA$ is the error in estimating the dictionary
  elements in \eqref{eqn:approx}.  
  \label{ass:exact-recovery}
\end{assumption}
\endgroup

\begin{theorem}[Exact recovery for bernoulli coefficients]
  Under the conditions of Theorem~\ref{thm:main-init}, and suppose, in
  addition Assumption~\ref{ass:exact-recovery} holds, then the output
  $\widehat{A}$ of Algorithm~\ref{algo:lasso} initialized with
  Algorithm~\ref{algo:init-dict} satisfies $\widehat{A} = A$ up to
  permutation of columns, with probability at least $1-3n^2\delta$.
  \label{thm:exact-recovery}
\end{theorem}

\paragraph{Remark: }
Assumption~\ref{ass:exact-recovery} for exact recovery places more
stringent conditions on the distribution of the coefficients and the
sparsity level $s$, compared to $(A4)$ for approximate recovery. While
for approximate recovery, we require $s=\order(d^{1/4}, r^{1/4})$, in
Assumption~\ref{ass:exact-recovery}, we require $s=\order(d^{1/5},
r^{1/6})$ for exact recovery. Note that the additional constraint on
sample complexity $n$ in Assumption~\ref{ass:exact-recovery} still has
the same scaling, and thus, $n=\order(r(\log r+\log d))$ suffices both
for approximate and exact recovery.

We also observe that the result of Theorem~\ref{thm:exact-recovery}
relies on Algorithm~\ref{algo:init-dict} as the initialization
procedure, but in principle we can also use a different approximate
recovery procedure to initialize Algorithm~\ref{algo:lasso}. In
particular, a different initialization procedure with a better error
guarantee would also directly translate to better recovery properties
in the second step, in terms of the assumptions relating $s$ to $r$
and $d$. Understanding these issues appears to be an interesting
direction for future research.


\section{Proofs of main results}
\label{sec:proofs}

In this section we will present the proofs of our main results,
Theorems~\ref{thm:main-init}
and~\ref{thm:exact-recovery}. We will start by presenting a
host of useful lemmas, and sketch out how they fit together to yield
the main results before moving on to the proofs.

\subsection{Correlation graph properties}
\label{sec:corrgraph}

In this section we will present some useful properties of the
correlation graph $G_{\corr(\rho)}$ described in
Section~\ref{sec:overview}. Recall that $G_{\corr(\rho)}$, where the
nodes are samples $\{y_1, y_2, \ldots y_n\}$ and an edge $(y_i,
y_j)\in G_{\corr(\rho)}$ implies that $|\inner{y_i, y_j}| > \rho$, for
some $\rho>0$. This is employed by Algorithm~\ref{algo:init-dict} as a
proxy for identifying samples which have common dictionary
elements. We now make this connection concrete in the next few
lemmas. For this we also recall our notation $\nbd_B(y)$ which is the
neighborhood of a sample $y$ in the coefficient bipartite graph (see
Figure~\ref{fig:bipartite}), that is, the set of dictionary elements
that combine to yield $y$.


\begin{lemma}[Correlation graph]\label{lemma:corrgraph}
Under the incoherence assumption $(A2)$ and the threshold $\rho$ in
assumption $(A7)$, the following is true for the edges in the
correlation graph $G_{\corr(\rho)}$:

\begin{align}| \nbd_B(y_k)\cap\nbd_B( y_l)| =1 &\Rightarrow (y_k,y_l)
  \in G_{\corr(\rho)}, \quad \forall\, i\in [r], \label{eqn:corr1}\\
  (y_k,y_l) \in G_{\corr(\rho)} &\Rightarrow | \nbd_B(y_k)\cap\nbd_B(
  y_l)| \geq 1,\label{eqn:corr2}
\end{align}
for all $k,l\in \{1,2,,\ldots, n\}, k\neq l$.
\end{lemma}

Lemma~\ref{lemma:corrgraph} suggests that nodes which intersect in
\emph{exactly one} dictionary element are special, in that they are
guaranteed to have an edge between them in $G_{\corr(\rho)}$. Our next
lemma works towards establishing something even stronger. We will next
establish that there are large cliques in the correlation graph where
any two samples in the clique intersect in the same unique dictionary
element. In order to state the lemma, we need some additional
notation.

For each dictionary element $a_i$, consider a set of
samples\footnote{Note that such a set need not be unique.} $\{y_k, k
\in S\}$, for some $S\subset \{1,2,\ldots, n\}$, such that they only
have $a_i$ in common, and denote such a set by $\Cc_i$ i.e.

\beq
\label{eqn:cc}
\Cc_i := \{ y_k, k\in S: \nbd_B(y_k)\cap
\nbd_B(y_l)= \{a_i,\} \,\, \forall\,k,l\in S\}.
\eeq
Lemma~\ref{lemma:corrgraph} implies that in the correlation graph, the
set of nodes in $\Cc_i$ form a clique (not necessarily maximal), for
each $i \in \{1,2,\ldots, r\}$, as shown in
Figure~\ref{fig:clustergraph}. The above implication can be exploited
for recovery of dictionary elements: if we find the set $\Cc_i$, then
we can hope to recover the element $a_i$, since that is the only
element in common to the samples in $\Cc_i$.

For ease of stating the next lemma, we further define two shorthand
notations.

\beq
\uniqint(y_i, y_j) :=\{(y_i, y_j) \in G_{\corr(\rho)} \quad \mbox{and}
\quad |\nbd_B(y_{i}) \cap \nbd_B(y_{j})|=1\},
\label{eqn:uniqint}
\eeq
Intuitively, the samples satisfying $\uniqint(y_i, y_j)$ are
guaranteed to have an edge between them by
Lemma~\ref{lemma:corrgraph}. In order to guarantee large cliques, we
will also need to measure the number of triangles in
$G_{\corr(\rho)}$.

In order to do this, given anchor samples $y_{i^*}$ and $y_{j^*}$ have
a unique intersection, we now bound the probability that a randomly
chosen sample $y_i$, among the neighborhood set of $y_{i^*}$ and
$y_{j^*}$ in the correlation graph also has a unique intersection. Now
define unique intersection event for a new sample $y_i$ with respect
to anchor samples $y_{i^*}$ and $y_{j^*}$ as follows

\beq
\uniqint(y_i; y_{i^*}, y_{j^*}):= \left\{\nbd_B(y_{i}) \cap
\nbd_B(y_{i^*})= \nbd_B(y_{i}) \cap \nbd_B(y_{j^*})=\{a_{k}\}\right\},
\eeq
where $a_k=\nbd_B(y_{i^*}) \cap \nbd_B(y_{j^*})\}$ is the unique
intersection of the anchor samples $y_{i^*}$ and $y_{j^*}$. In other
words, $\uniqint(y_i; y_{i^*}, y_{j^*})$ indicates the event that the
pairwise intersections of the new sample $y_i$ with each of the
anchors $y_{i^*}$ and $y_{j^*}$ is unique and equal to the unique
intersection of $y_{i^*}$ and $y_{j^*}$.

\begin{lemma}[Formation of clique under good anchor samples]
  \begin{align*}
    &\Pbb\left[\uniqint(y_i; y_{i^*}, y_{j^*})\,\big|\,
      \uniqint(y_{i^*}, y_{j^*}),\mbox{ and } (y_i, y_{i^*}), (y_i,
      y_{j^*})\in G_{\corr(\rho)}\right]\\
    &\geq 1- \frac{s^3}{r}.
  \end{align*}
\label{lemma:good-clique}
\end{lemma}

Lemma~\ref{lemma:good-clique} is crucial for our algorithm. It
guarantees that given a pair of good anchor elements---one satisfying
unique intersection property---a large fraction of their neighrbors
also contain this common dictionary element. Some further arguments
can then be made to establish that a large fraction of the neighbors
of $y_{i^*}$ and $y_{j^*}$ also have edges amongst themselves and
hence form cliques as defined in Equation~\ref{eqn:cc}.

\subsection{Correctness of Procedure~\ref{procedure:uniqueint}}
\label{sec:uniqueint}

A key component in our analysis is the correctness of
Procedure~\ref{procedure:uniqueint}. As we saw in the previous lemmas,
it is crucial for a chosen pair of anchor elements to have a unique
intersection in order to use them for identifying large cliques
$\Cc_i$ in $G_{\corr(\rho)}$. Procedure~\ref{procedure:uniqueint} plays
a crucial role by providing a verifiable test for whether a pair of
anchor elements have a unique intersection or not. Our next two lemmas
help us establish that this test is sound with high probability. We
first show that two neighbors of a bad anchor pair do not have an edge
amongst them with high probability.

Denote the event

\[\Delta(y_i, y_j, y_k):= \{ (y_i, y_j), (y_j, y_k), (y_i, y_k) \in
G_{\corr(\rho)}\},\]
i.e., the samples $y_i, y_j, y_k$ form a triangle in the correlation
graph.

\begin{lemma}[Detection of bad anchor samples]
  \label{lem:detect-bad-anchor}
  For randomly chosen samples $y_i, y_j$
  \[ \Pbb\left[ (y_i, y_j)\notin G_{\corr(\rho)} \mid \Delta(y_i, y_{i^*},
    y_{j^*}), \Delta(y_j, y_{i^*}, y_{j^*}), \neg\uniqint(y_{i^*},
    y_{j^*})\right]>\frac{1}{16}.\]
\end{lemma}

Intuitively, this means that the number of sets $S_i$ which will be
edges in $G_{\corr(\rho)}$ is rather small for an anchor pair with
multiple dictionary elements in common. In order for correctness of
the procedure, we will in fact need this number to be substantially
smaller than that for a good anchor pair. This is indeed the case as
we next establish.

\begin{lemma}[Detection of good anchor samples]
  \label{lem:detect-good-anchor}
  For randomly chosen samples  $y_i, y_j$

  \[ \Pbb\left[ (y_i, y_j)\notin G_{\corr(\rho)} \mid \Delta(y_i,
    y_{i^*}, y_{j^*}), \Delta(y_j, y_{i^*}, y_{j^*}),
    \uniqint(y_{i^*}, y_{j^*})\right] \leq \frac{24 s^3}{r}.\]
\end{lemma}

Combining the above two lemmas, the correctness of
Procedure~\ref{procedure:uniqueint} naturally follows.

\begin{proposition}[Correctness of Procedure~\ref{procedure:uniqueint}]
\label{prop:detect-unique-int}
  Suppose $(y_{i^*}, y_{j^*}) \in G_{\corr(\rho)}$. Suppose that $s^3
  \leq r/1536$ and $\gamma \leq 1/64$. Then Algorithm~\ref{algo:init-dict}
  returns the value of $\uniqint(y_{i^*}, y_{j^*})$ correctly with
  probability greater than $1 - 2\exp(-\gamma^2 \subsamp)$.
\end{proposition}

\subsection{Proof of Theorem~\ref{thm:main-init}}
\label{sec:svd}

In this section we will put all the pieces together and establish
Theorem~\ref{thm:main-init}. We start by establishing that given a
pair of good anchor elements, the SVD step in
Algorithm~\ref{algo:init-dict} approximately recovers the unique
dictionary element in the intersection of the two anchors.

\begin{proposition}[Accuracy of SVD]\label{prop:accuracy-SVD}
  Consider anchor samples $y_{i^*}$ and $y_{j^*}$ such that
  $\uniqint(y_{i^*}, y_{j^*})$ is satisfied, and wlog, let
  $\nbd_B(y_{i^*})\cap\nbd_B(y_{j^*}) = \{a_1\}$.  Recall the
  definition of $\hS$~\eqref{eqn:good-anchor-svd-sets}, and further
  define $\hL:=\sum_{i\in \hS} y_i y_i^\top$ and $\widehat{n}=|\hS|$. If
  $\widehat{a}$ is the top singular vector of $\hL$, then there exists
  a universal constant $c$ such that we have:

\begin{align*}
  \min_{z\in \set{-1,1}} \twonorm{\widehat{a}-za_1}^2 <
  \frac{32sM^2}{m^2}\left(\frac{\mu_1}{\sqrt{ds}}+ \frac{\mu_1^2}{d} + \frac{s^3}{r}
  + \alpha^2 + \frac{\alpha}{\sqrt{s}}\right),
\end{align*}
with probability greater than $1 - d\exp\left(-c\alpha^2 \widehat{n}\right)$
for $\alpha < 1/20$.
\end{proposition}

Given the above proposition, the proof of Theorem~\ref{thm:main-init}
is relatively straightforward. Indeed, the key missing piece is the
dependence on the random quantity $|\hS|$ in the error probability in
Proposition~\ref{prop:accuracy-SVD}. We now present the proof.

\bprfof{Theorem~\ref{thm:main-init}}

Consider a particular iteration of Algorithm~\ref{algo:init-dict}.
Procedure~\ref{procedure:uniqueint} returns $\uniqint(y_{i^*},
y_{j^*})$ with probability greater than $1 -
2\exp(-\gamma^2\abs{\widehat{S}}/2)$.  If $\neg \uniqint(y_{i^*},
y_{j^*})$, then Algorithm~\ref{algo:init-dict} proceeds to the next
iteration. Consider the case of $\uniqint(y_{i^*}, y_{j^*})$ and
suppose $\nbd_B(y_{i^*}) \cap \nbd_B(y_{j^*}) = \set{a_l}$. Using
Proposition~\ref{prop:accuracy-SVD}, with probability greater than $1
- d\exp\left(-c\alpha^2\abs{\widehat{S}}\right)$, we have:
\begin{align*}
  \twonorm{a_l - \widehat{a}}^2 <
  \frac{32sM^2}{m^2}\left(\frac{\mu_1}{\sqrt{ds}}+ \frac{\mu_1^2}{d} + \frac{s^3}{r}
  + \alpha^2 + \frac{\alpha}{\sqrt{s}}\right).
\end{align*}
Using Lemma~\ref{lem:size-Shat} and Lemma~\ref{lemma:corrgraph}, we
see that $\abs{\widehat{S}} \geq \frac{ns}{4r}$ with probability
greater than $1 - \exp\left(\frac{-ns}{16r}\right)$.  Using a union
bound over all the iterations (which are at most $n^2$), the above
claims hold for all iterations with probability greater than $1 - n^2d
\exp\left(\frac{-c\alpha^2ns}{r}\right) - 2n^2
\exp\left(\frac{-\gamma^2 n s}{8r}\right) -
n^2\exp\left(\frac{-ns}{16r}\right)$.

Using Lemma~\ref{lem:size-Shat} and Lemma~\ref{lemma:corrgraph}, with
probability greater than $1 - r\exp\left(\frac{-ns}{64r}\right)$, for
every $l \in [r]$, there are at least $\frac{ns}{8r}$ pairs
$(i^*,j^*)$ such that $\nbd_B(y_{i^*}) \cap \nbd_B(y_{j^*}) =
\set{a_l}$ and $(i^*,j^*) \in G_{\corr(\rho)}$.  Lines 9-11 of the
algorithm then ensure that there is a unique copy of the approximation
to $a_l$ dictionary element. Using a union bound now gives the result.

\eprfof

\subsection{Analysis of post-processing step}
\label{sec:lasso}

In this section, we will show how to clean up the approximate recovery
of the previous section and obtain exact recovery of the dictionary
under Assumption~\ref{ass:exact-recovery}. We start by setting up the
problem as that of sparse estimation with deterministic noise and
describing some guarantees in a general setup. We then specialize
these to the assumptions of our problem and present the proof of
Theorem~\ref{thm:exact-recovery}.

\subsubsection{Lasso with determinstic noise}

Recalling the model~\eqref{eqn:dict}, we see that each observation
$y_i$ is generated according to the linear model

\begin{equation*}
  y_i = Ax_i, \quad \mbox{for $i = 1,2,\ldots, n$},
\end{equation*}
where $x_i$ is a $\spindex$-sparse vector in $r$ dimensions. If we
knew the dictionary $A$, then this is the usual sparse linear
system. Given the knowledge of an approximate dictionary $\Ainit$
however, we can rewrite the system as

\begin{equation}
  y_i = \Ainit x_i + \underbrace{(A - \Ainit)x_i}_{w_i},
  \label{eqn:approx-linear}
\end{equation}
where $W \in \R^{d\times n}$ is the error matrix. Note that the errors
in $W$ are not zero mean, or even independent of $\Ainit$ unlike
typical statistical settings. Under our initialization, however, they
are bounded, which we establish subsequently. For the remainder of
this section, we assume the following facts about $\Ainit$. Note that
this is not an assumption about the model, but a condition on the
output of Algorithm~\ref{algo:init-dict}, which will be proved in the
next section.

\begingroup
\def\theassumption{C1}
\begin{assumption}[Approximate initialization] Assume that $\Ainit$ is
  an approximately correct initialization for $A$, meaning the
  following hold:
\item \textbf{RIP:} The $2\spindex$-RIP constant of the matrix
  $\Ainit$, $\delta_{2\spindex} < \frac{1}{7}$. That is, for every $S
  \subseteq \{1,2,\ldots, r\}$ with $|S| \leq 2\spindex$, the smallest
  and largest singular values, $\singmin$ and $\singmax$ respectively
  of the $d\times |S|$ matrix $\Ainit_S$ satisfy:
$$ \frac{6}{7} < \singmin < \singmax < \frac{8}{7}.$$
\item \textbf{Bounded error:} $\norm{\ainit_i - a_i}_2 \leq \errA$
  for all $i = 1,2,\ldots, r$.
  \label{ass:approx-init}
\end{assumption}
\endgroup

Under these general assumptions, we can provide a guarantee on the
error incurred in~\eqref{eqn:lasso} in  step (2) of
Algorithm~\ref{algo:lasso}. While this result has been obtained in
many contexts by various authors, we use the following precise form from
Candes~\cite{Candes2008}.

\begin{theorem}[Theorem 1.2 from Candes~\cite{Candes2008}]
  Suppose $y_i$ is generated according to the linear
  model~\eqref{eqn:approx-linear}, where $x_i$ is $s$-sparse and assume
  that $\delta_{2s} \leq \sqrt{2}-1$. Then the
  solution to Equation~\eqref{eqn:lasso} obeys the following, for a universal constant
  $C_1$,
  \begin{equation*}
    \norm{\widehat{x}_i - x_i}_2 \leq C_1\norm{w_i}_2.
  \end{equation*}
  In particular, $C_1 = 8.5$ suffices for $\delta_{2s} \leq 0.2$.
  \label{thm:candes}
\end{theorem}

\subsubsection{Proof of Theorem~\ref{thm:exact-recovery}}

In order to prove Theorem~\ref{thm:exact-recovery}, we first establish
that under our assumptions, the coefficients $x_i$ are exactly
recovered in Equation~\eqref{eqn:lasso}. Once this is established,
Theorem~\ref{thm:exact-recovery} follows in a straightforward
manner. We start with a useful proposition.

\begin{proposition}
  Under conditions of Theorem~\ref{thm:main-init}, assume further that
  $\errA \leq 1/(20s)$ for the dictionary returned in
  Algorithm~\ref{algo:init-dict}. Then Algorithm~\ref{algo:lasso}
  guarantees that $\widehat{x_i} = x_i$ for all $i = 1,2,\ldots, n$.
  \label{prop:exact}
\end{proposition}

\bprf We would like to use Theorem~\ref{thm:candes} to show that we
recover the coefficients $x_i$ correctly in the lasso
step~\eqref{eqn:lasso} of Algorithm~\ref{algo:lasso}. In order to do
this, we first need to verify Assumption~\ref{ass:approx-init} for the
dictionary returned by Algorithm~\ref{algo:init-dict}, and then obtain
bounds on the quantity $\norm{w_1}_2$. We start with the former.

Consider any $2\spindex$-sparse subset $S$ of $[r]$. We have:

\begin{align*}
  \singmin(\Ainit_S) &\geq \singmin(A_S) - \norm{A_S - \Ainit_S}_2
  \stackrel{(\zeta_1)}{\geq} 1 - \frac{2\mu_0 \spindex}{\sqrt{d}} -
  \frob{A_S - \Ainit_S} \quad \mbox{ and,}
  \\
  \singmax(\Ainit_S) &\leq \singmax(A_S) + \norm{A_S - \Ainit_S}_2
  \stackrel{(\zeta_2)}{\leq} 1 + \frac{2\mu_0 \spindex}{\sqrt{d}} +
  \frob{A_S - \Ainit_S},
\end{align*}
where $\zeta_1$ and $\zeta_2$ follow from Lemma~\ref{lem:RIP-A} in
Appendix~\ref{app:aux}. Since $A_S$ is a $d \times 2s$ matrix, it
satisfies that $\frob{A_S - \Ainit_S} \leq \sqrt{2s\epsilon_A}$. Given
the assumption $\epsilon_A \leq 1/(20s)$, it immediately follows that
the minimum and maximum singular values of $\Ainit_s$ are at least
$6/7$ and $8/7$ respectively, so that we obtain $\delta_{2s} = 1/7 <
0.2$.

This shows that $\Ainit$ satisfies
Assumption~\ref{ass:approx-init}. Next we bound the $\ell_2$ norm of
the noise vector $w_i$. Again bounding the frobenius norm of the error
in the dictionary in the same way as above, we obtain

\begin{align*}
  \norm{w_i}_2 \leq \norm{(A-\Ainit)_{S_i}}_2 \norm{x_i}_2 \leq
  \frob{(A-\Ainit)_{S_i}} \sqrt{ \spindex} \leq \spindex \errA,
\end{align*}
where $S_i$ is the support of $x_i$. Consequently, we obtain from
Theorem~\ref{thm:candes} that the output $\widehat{x}_i$ of
Equation~\ref{eqn:lasso} satisfies

\begin{equation}
  \norm{\widehat{x}_i - x_i}_2 \leq C_1\, s\errA \leq
  9s\errA \leq 9/20.
  \label{eqn:lasso-final-ell2}
\end{equation}

We now observe that an $\ell_2$ error guarantee is also an
$\ell_\infty$ error guarantee. Recall that by the model assumption,
each non-zero coefficient of $X$ has an absolute value of $1$. Since
Equation~\eqref{eqn:lasso-final-ell2} guarantees that the $\ell_2$
error guarantee is no larger than $1/2$, all the coefficients will be
uniquely recovered and hence $\widehat{x}_i = x_i$.

\eprf

\bprfof{Theorem~\ref{thm:exact-recovery}}

We are now ready to provide our proof of exact recovery. Based on
Proposition~\ref{prop:exact}, we only need to verify two things. First
is that the initialization $\Ainit$ satisfies $\errA < 1/(20s)$ and
the second is that the linear system $Y = AX$ is well-posed when we
solve for $A$. In order to verify the former, we observe that our
additional conditions in Assumption~\ref{ass:exact-recovery} guarantee
that

\begin{align*}
  32s\left( \frac{\mu_1}{\sqrt{ds}} + \frac{\mu_1^2}{d}\right) &\leq
  \frac{1}{1200s^2},\\
  \frac{32s^4}{r} &\leq \frac{1}{1200s^2}, \quad \mbox{and}\\
  32s\left( \alpha^2 + \frac{\alpha}{\sqrt{s}} \right) &\leq
    \frac{1}{1200s^2}.
\end{align*}

Hence we obtain from Theorem~\ref{thm:main-init} that with probability
at least $1 - n^2d\exp\left(\frac{-c\alpha^2ns}{r}\right) -
4\max(n^2,r) \exp\left(\frac{-n s}{32768r}\right)$, $\errA <
1/(20s)$. Hence, it only remains to verify that the linear system is
well-posed.

According to Lemma~\ref{lemma:X-covar-matrix} in
Appendix~\ref{app:aux}, the matrix $\expec{XX^T} = \frac{s}{r}
I_{r\times r}$ so that all of its singular values are equal to
$s/r$. We now appeal to
Theorem~\ref{thm:rand-matrix-spectralnorm-vershynin} with $W = X$, $d
= r$ and $u = \sqrt{s}$. Then we obtain for any $t > 0$ with
probability at least $1 - r\exp(-ct^2)$

 \begin{align*}
  \singmin(XX^T) \geq \frac{ns}{r} - n \max\left\{
  \sqrt{\frac{s}{r}} \delta, \delta^2 \right\},
\end{align*}
where $\delta = t\sqrt{s/n}$. Substituting the value of $\delta$, we
obtain the lower bound

\begin{align*}
  \singmin(XX^T) &\geq \frac{ns}{r} - n \max\left\{
  \sqrt{\frac{s}{r}} t\sqrt{\frac{s}{n}}, t^2\frac{s}{n} \right\}\\
  &\geq \frac{ns}{r} \left(1 - t\sqrt{\frac{r}{n}} - \frac{t^2r}{n}
  \right)\\
  &= \frac{ns}{4r},
\end{align*}
for $t = \sqrt{n/(4r)}$. This means that the linear system is
well-posed with probability at least $1 - r\exp(-cn/(4r))$.  Choosing
$c_0$ to now be $\min(c,1/32768)$ finishes the proof.
\eprfof

\section{Discussion and Conclusion}


In this paper, we proposed simple and tractable methods for dictionary
learning. We present a novel clustering-based approach which can
approximately recover the uknown overcomplete dictionary from
samples. We also analyzed a simple denoising strategy based on sparse
recovery algorithms for reconstructing the dictionary exactly under
some simplifying assumptions on the model. In particular, the second
step is not tied to the first step in any critical way, and more
sophisticated post-processing procedures have since been
developed. There is of course, also room for developing better
approximate recovery schemes, building on our work.

In the analysis of the clustering step, we provide guarantees when the
coefficient matrix is sparse and randomly drawn.  In principle, our
analysis can be extended to general sparse coefficient matrices and
can be cast as a higher-order expansion condition on the coefficient
bipartite graph. Similar (and yet not the same) expansion conditions
have appeared in other contexts involving learning of overcomplete
models. For instance, in~\cite{AnandkumarEtal:overcomplete13},
Anandkumar et. al. establish that under an expansion condition on the
topic-word matrix, unsupervised learning of the model is
possible. Here, the hidden topics correspond to dictionary elements,
and the observed words correspond to the samples in the dictionary
setting.

Finally, our work suggests some natural and interesting directions for
future research. While both the steps of our algorithm seem inherently
robust to noise, it remains important to quantify the recovery
properties when the observations are noisy in future work. Another
natural question is raised by the fact that we use only one step of
lasso and least squares for exact recovery. Indeed, the subsequent
work~\cite{AgarwalAJNT13} analyzes a generalization where we perform
multiple iterations of lasso followed by subsequent dictionary
estimation, and is able to exactly recover the dictionary under a much
broader set of conditions. Since our study was motivated by natural
applications of dictionary learning in signal processing and machine
learning, it would also be interesting to investigate how our provably
correct procedures perform compared to the popular heuristic methods.

\subsubsection*{Acknowledgements}

A. Agarwal thanks Yonina Eldar for suggesting the problem to
him. A. Anandkumar is supported in part by Microsoft Faculty
Fellowship, NSF Career award CCF-1254106, NSF Award CCF-1219234, and
ARO YIP Award W911NF-13-1-0084. P. Netrapalli thanks Yash Deshpande for
helpful discussions. The authors thank Matus Telgarsky for
suggesting Lemma~\ref{lemma:binom-lower} and thank Sham Kakade and
Dean Foster for initial discussions.

\newpage

\appendix

\section{Proofs for clustering analysis}

In this section we will provide the proofs of many of the Lemmas along
with some auxilliary results in
Sections~\ref{sec:corrgraph}-~\ref{sec:svd}. Some of the more
technical results that are required will be deferred to
Appendix~\ref{app:aux}.


\subsection{Proofs of correlation graph properties}

We start by proving Lemmas~\ref{lemma:corrgraph}
and~\ref{lemma:good-clique} in Section~\ref{sec:corrgraph}.

\bprfof{Lemma~\ref{lemma:corrgraph}}

We first prove \eqref{eqn:corr2} via contradiction. Suppose
$\nbd_B(y_k) \cap \nbd_B(y_l)=\emptyset$, we then have

\begin{align*}
|\inner{y_k ,y_l}| &= |\sum_{i,j} x_{ik} x_{jl} \inner{a_i, a_j}| \leq
\sum_{i,j} |x_{ik} x_{jl} \inner{a_i, a_j}| \\
& \leq |\nbd_B(y_k)|\cdot |\nbd_B(y_l)|\cdot \max_{i,j,k,l} |x_{ik}
x_{jl}|\cdot\max_{i\neq j}| \inner{a_i, a_j}|  \leq
\frac{s^2 M^2 \mu_0}{\sqrt{d}}
\end{align*}
For \eqref{eqn:corr1}, let $\{a_{i^*}\}= \nbd_B(y_k)\cap \nbd_B(y_l)$
\begin{align*}
|\inner{y_k ,y_l}| &= |\sum_{i,j} x_{ik} x_{jl} \inner{a_i, a_j}| \geq
|x_{i^*k} x_{i^*l}| \inner{a_{i^*}, a_{i^*}}- \sum_{i\neq j} |x_{ik} x_{jl}
\inner{a_i, a_j}|\\
& \geq m^2 - \frac{s^2 M^2 \mu_0}{\sqrt{d}},
\end{align*}
using the above analysis. The claims now follow from the setting of
$\rho$.
\eprfof

We next establish Lemma~\ref{lemma:good-clique}.

\bprfof{Lemma~\ref{lemma:good-clique}}
Define the event

\[ \Ac:= \{|\nbd_B(y_i)\cap\nbd_B(y_{i^*})|\geq 1 \}\cap
\{|\nbd_B(y_i)\cap\nbd_B(y_{j^*})|\geq 1 \}.\]
From Lemma~\ref{lemma:corrgraph}, we have that

\begin{align*}
  &\Pbb\left[\uniqint(y_i; y_{i^*}, y_{j^*})\,\big|\, \uniqint(y_{i^*},
    y_{j^*}),\mbox{ and } (y_i, y_{i^*}), (y_i, y_{j^*})\in
    G_{\corr(\rho)}\right]\\
  &\geq\Pbb\left[\uniqint(y_i; y_{i^*}, y_{j^*})\,\big|\,
    \uniqint(y_{i^*}, y_{j^*}),\Ac\right]
\end{align*}
In order to lower bound $\Pbb\left[\uniqint(y_i; y_{i^*},
  y_{j^*})\,\big|\, \uniqint(y_{i^*}, y_{j^*}),\Ac\right]$, we instead
upper bound the probability of the complementary event
$\Pbb\left[\lnot \uniqint(y_i; y_{i^*}, y_{j^*})\,\big|\,
  \uniqint(y_{i^*}, y_{j^*}),\Ac\right]$

In order to do so, we first bound the following

\begin{align}
\Pbb\left[\Ac\,\big|\, \uniqint(y_{i^*}, y_{j^*})\right]\geq
\frac{s}{r},
\label{eqn:cliquep-bound1}
\end{align}
since $\Ac$ holds when the unique element in $\nbd_B(y_{i^*})\cap
\nbd_B(y_{j^*})$ is chosen and its probability is $s/r$. We also have

\begin{align*}
  \Pbb\left[\lnot\uniqint(y_i; y_{i^*}, y_{j^*})\cap \Ac\,\big|\,
  \uniqint(y_{i^*}, y_{j^*})\right] \leq
  \frac{(s-1)^2 \binom{r-3}{s-2}}{\binom{r}{s}},
\end{align*}
since for $\lnot\uniqint(y_i; y_{i^*}, y_{j^*})$ to hold, we need to
choose at least one of the $s-1$ elements in
$\nbd_{B}(y_{i^*})/\nbd_{B}(y_{j^*})$, and similarly one from the $s-1$
elements of $\nbd_{B}(y_{j^*})/\nbd_{B}(y_{i^*})$. The rest of the $s-2$
elements can be picked arbitrarily from the $r-3$ dictionary atoms
that remain after excluding the two already picked and the unique
intersection $\nbd_{B}(y_{j^*}) \cap \nbd_{B}(y_{i^*})$.

It is easy to check that

\begin{align}
  \frac{(s-1)^2 \binom{r-3}{s-2}}{\binom{r}{s}} &= \frac{(s-1)^2
    (r-s)s(s-1)}{r(r-1)(r-2)} \nonumber\\
  &\leq \frac{s^4}{r^2}.
  \label{eqn:cliquep-bound2}
\end{align}
Taking the ratio of the two bounds in~\eqref{eqn:cliquep-bound1}
and~\eqref{eqn:cliquep-bound2} completes the proof.
\eprfof


\subsection{Proofs of Lemmas~\ref{lem:detect-bad-anchor}
  and~\ref{lem:detect-good-anchor}}

We now prove the two lemmas that are crucial to establishing the
correctness of Prcedure~\ref{procedure:uniqueint}.

\bprfof{Lemma \ref{lem:detect-bad-anchor}}
Let $\Ac_1$ and $\Ac_2$ denote the following events:
\begin{align}
  \Ac_1:=& \{|\nbd_B(y_i)\cap \nbd_B(y_{i^*})|\geq 1\}\cap
  \{|\nbd_B(y_i)\cap \nbd_B(y_{j^*})|\geq 1\} \nonumber \\
  &\cap \{|\nbd_B(y_j)\cap \nbd_B(y_{i^*})|\geq 1\} \cap
  \{|\nbd_B(y_j)\cap \nbd_B(y_{j^*})|\geq 1\} \nonumber\\
  \Ac_2:=& \{|\nbd_B(y_i)\cap \nbd_B(y_{i^*})|= 1\}\cap
  \{|\nbd_B(y_i)\cap \nbd_B(y_{j^*})|= 1\} \nonumber \\
  &\cap \{|\nbd_B(y_j)\cap \nbd_B(y_{i^*})|= 1\} \cap
  \{|\nbd_B(y_j)\cap \nbd_B(y_{j^*})|= 1\}
  \label{eqn:unique-int-events}
\end{align}
In words, both $y_i$ and $y_j$  have at least dictionary element in
common with each of $y_{i^*}$ and $y_{j^*}$ under the event $\Ac_1$,
while the number of common elements is \emph{exactly one} under the
event $\Ac_2$. We have

\begin{align}
  & \Pbb\left[ (y_i, y_j)\notin G_{\corr(\rho)} \mid \Delta(y_i, y_{i^*},
    y_{j^*}), \Delta(y_j, y_{i^*}, y_{j^*}), \neg\uniqint(y_{i^*},
    y_{j^*})\right] \nonumber \\
  &\stackrel{(a)}{=} \Pbb\left[ (y_i, y_j)\notin G_{\corr(\rho)} \mid
    \Ac_1, \Delta(y_i, y_{i^*}, y_{j^*}), \Delta(y_j, y_{i^*},
    y_{j^*}), \neg\uniqint(y_{i^*}, y_{j^*})\right] \nonumber \\
  & = \Pbb\left[ (y_i, y_j)\notin G_{\corr(\rho)}, \Delta(y_j,
    y_{i^*}, y_{j^*}) \mid\Ac_1,  \Delta(y_i, y_{i^*}, y_{j^*}),
    \Delta(y_j, y_{i^*}, y_{j^*}), \neg\uniqint(y_{i^*},
    y_{j^*})\right] \nonumber \\
  &\geq \Pbb\left[ (y_i, y_j)\notin G_{\corr(\rho)}, \Delta(y_i,
    y_{i^*}, y_{j^*}), \Delta(y_j, y_{i^*}, y_{j^*}) \mid
    \Ac_1, \neg\uniqint(y_{i^*}, y_{j^*}), (y_{i^*}, y_{j^*}) \in
    G_{\corr(\rho)}\right] \nonumber \\
  &\stackrel{(b)}{\geq} \Pbb\left[ (y_i, y_j)\notin G_{\corr(\rho)},
    \Ac_2 \mid \Ac_1, \neg\uniqint(y_{i^*}, y_{j^*}),(y_{i^*},
    y_{j^*}) \in G_{\corr(\rho)}\right] \nonumber
  \\
  &\stackrel{(c)}{\geq} \Pbb\left[ \{\nbd_B(y_i)\cap\nbd_B( y_j)
    =\emptyset\} \cap \Ac_2 \mid \Ac_1, \neg\uniqint(y_{i^*},
    y_{j^*}), (y_{i^*}, y_{j^*}) \in G_{\corr(\rho)}\right],
  \label{eqn:bad-prob-to-lb}
\end{align}
where the inequalities $(a)$, $(b)$ and $(c)$ follow from
Lemma~\ref{lemma:corrgraph}. We will now work on lower bounding this
resulting probability.

We first lower bound the numerator in writing the above conditional
probability as the ratio of a joint to marginal probability. We begin
by noting that

\begin{align*}
  & \Pbb\left[ \{ \nbd_B(y_i)\cap\nbd_B( y_j) =\emptyset\}\cap \Ac_2
    \cap \Ac_1 \mid  \neg\uniqint(y_{i^*}, y_{j^*}, (y_{i^*}, y_{j^*})
    \in G_{\corr(\rho)})\right]\\
  &= \Pbb\left[ \{ \nbd_B(y_i)\cap\nbd_B( y_j) =\emptyset\}\cap \Ac_2
    \mid  \neg\uniqint(y_{i^*}, y_{j^*}), (y_{i^*}, y_{j^*}) \in
    G_{\corr(\rho)}\right]
\end{align*}

Let us define $\widehat{l}= |\nbd_B(y_{i^*})\cup\nbd_B(y_{j^*})|\in [s,2s]$ and
$l = |\nbd_B(y_{i^*})\cap\nbd_B(y_{j^*})|\geq 2$\footnote{the
  intersection is at least 1 by Lemma~\ref{lemma:corrgraph}}. The
event in the probability above, that is $\Ac_2$ holds while $y_i$ and
$y_j$ do not share a dictionary element, can be arranged by choosing
two of the $l$ elements, and assigning a unique element to each $y_i$
and $y_j$. Similarly the remaining elements can be chosen outside
$\nbd_B(y_{i^*})\cup\nbd_B(y_{j^*})$ in a non-overlapping manner: for
$y_i$ assign $s-1$ elements among $r-\widehat{l}$ elements, and then for $y_j$
assign from remaining $r-\widehat{l}-s+1$ elements. This logic yields the
following lower bound on the probability

\begin{align*}
  & \Pbb\left[ \{ \nbd_B(y_i)\cap\nbd_B( y_j) =\emptyset\}\cap \Ac_2
    \mid  \neg\uniqint(y_{i^*}, y_{j^*})\right]\\
  &\geq \frac{ 2 \binom{l}{2} \binom{r-\widehat{l}}{s-1}
    \binom{r-\widehat{l}-s+1}{s-1}}{\binom{r}{s}^2}\geq \frac{ 2 \binom{l}{2}
    \binom{r-2s}{s-1} \binom{r-3s+1}{s-1}}{\binom{r}{s}^2},
\end{align*}
where the second inequality uses $\widehat{l} \leq 2s$. Now with some
straightforward algebra, we can further lower bound this expression as

\begin{align*}
  & \Pbb\left[ \{ \nbd_B(y_i)\cap\nbd_B( y_j) =\emptyset\}\cap \Ac_2 \mid
    \neg\uniqint(y_{i^*}, y_{j^*})\right]\\
  &\geq \frac{s^2(l-1)^2}{r^2}\left(1 - \frac{3s-3}{r-s}\right)^{s-1}
  \left(1 - \frac{2s-1}{r-s}\right)^{s-1}\\
  &\geq
  \frac{s^2(l-1)^2}{r^2}\left(1-\frac{3s}{r-s}\right)^s\left(1-\frac{2s}{r-s}\right)^s.
\end{align*}
Now we invoke Lemma~\ref{lemma:binom-lower} to further lower bound the
RHS and obtain

\begin{align*}
  & \Pbb\left[ \{ \nbd_B(y_i)\cap\nbd_B( y_j) =\emptyset\}\cap \Ac_2 \mid
    \neg\uniqint(y_{i^*}, y_{j^*})\right]\\
  &\geq
  \frac{s^2(l-1)^2}{r^2}\exp\left(-\frac{3s^2}{r-s}\right)\exp\left(-\frac{2s^2}{r-s}\right)
  \geq \frac{s^2(l-1)^2}{r^2} \left(1 - \frac{10s^2}{r-s} \right)\\
  &\geq \frac{s^2(l-1)^2}{2r^2},
\end{align*}
where the final inequality holds since $s^2 \leq r/40$.

In order to lower bound the conditional probability in
Equation~\ref{eqn:bad-prob-to-lb}, we need to further upper bound the
marginal probability in the denominator. To this end, we observe that
we have to upper bound $\Pbb\left[\Ac_1|\neg\uniqint(y_{i^*},
  y_{j^*})\right]$. Now conditioned on $\neg\uniqint(y_{i^*},
y_{j^*})$, for each $y_i$ and $y_j$, $\Ac_1$ can be satisfied in two
ways: choose at least one element from $l$ elements in
$\nbd_B(y_{i^*})\cap\nbd_B(y_{j^*})$ or choose at least two elements
from $m-l$ elements in $\nbd_B(y_{i^*})\cup\nbd_B(y_{j^*})$. Making
this precise, we obtain

\begin{align*}
  \Pbb\left[\Ac_1\mid\neg\uniqint(y_{i^*}, y_{j^*})\right]
  &\leq \left(\frac{ls}{r} +
  \frac{(\widehat{l}-l)^2\binom{r-2}{s-2}}{\binom{r}{s}}\right)^2 \\
  &\leq \left(\frac{ls}{r} + \frac{s^2(\widehat{l}-l)^2}{(r-1)^2}\right)^2 \\
  &\leq \left(\frac{ls}{r} + \frac{s^2(2s-2)^2}{(r-1)^2}\right)^2 \\
  &\leq \frac{2l^2s^2}{r^2}, \;(\mbox{since } 4s^3 < r-1)
\end{align*}
The result follows by using the fact that $l \geq 2$.
\eprfof


The proof of Lemma~\ref{lem:detect-good-anchor} is similar, but
involves controlling slightly different events.

\bprfof{Lemma \ref{lem:detect-good-anchor}}

We will establish the lemma by lower bounding the probability of the
complementary event. We recall the events $\Ac_1$ and $\Ac_2$ defined
in Equation~\ref{eqn:unique-int-events} in the proof of
Lemma~\ref{lem:detect-bad-anchor}. We can mimick the initial arguments
in the proof of Lemma~\ref{lem:detect-bad-anchor} to conclude that

\begin{align*}
  & \Pbb\left[ (y_i, y_j)\in G_{\corr(\rho)} \mid \Delta(y_i, y_{i^*},
    y_{j^*}), \Delta(y_j, y_{i^*}, y_{j^*}), \uniqint(y_{i^*},
    y_{j^*})\right] \\
  &\geq \Pbb\left[ \uniqint(y_i,y_j) \cap \Ac_2 \mid \Ac_1,
    \uniqint(y_{i^*}, y_{j^*})\right],
\end{align*}
and we provide a lower bound for this. Once again, we express the
conditional probability as the ratio of a joint to a marginal and then
lower bound the numerator and upper bound the denominator. In the
numerator, we have the event

We have

\begin{align*}&
  \Pbb\left[ \uniqint(y_i,y_j) \cap \Ac_2 \cap \Ac_1 \mid
    \uniqint(y_{i^*}, y_{j^*})\right]\\
&= \Pbb\left[ \uniqint(y_i,y_j) \cap \Ac_2 \mid  \uniqint(y_{i^*},
  y_{j^*})\right]
\end{align*}

The event $\uniqint(y_i,y_j) \cap \Ac_2$ is guaranteed to occur if we
choose $y_i$ and $y_j$ so that they have the only element in
$\nbd_B(y_{i^*})\cap\nbd_B(y_{j^*})$ in common. This yields the lower
bound

\begin{align*}
  &\Pbb\left[ \uniqint(y_i,y_j) \cap \Ac_2 \cap \Ac_1 \mid
    \uniqint(y_{i^*}, y_{j^*})\right]\\
  &\geq \frac{\binom{r-2s+1}{s-1}\binom{r-3s+2}{s-1}}{\binom{r}{s}^2}.
\end{align*}
It is easy to further conclude that

\begin{align*}
  &\Pbb\left[ \uniqint(y_i,y_j) \cap \Ac_2 \cap \Ac_1 \mid
    \uniqint(y_{i^*}, y_{j^*})\right]\\
  &\geq \frac{s^2}{r^2} \left( 1 - \frac{3s-3}{r-s+1}\right)^{(s-1)}
  \left(1 - \frac{2s-2}{r-s+1} \right)^{s-1}\\
  &\geq \frac{s^2}{r^2} \exp(-5(s-1)^2/(r-s+1))\\
  &\geq \frac{s^2}{r^2}\left(1 - \frac{10s^2}{r-s} \right) \geq
  \frac{s^2}{r^2} \left( 1 - \frac{20s^2}{r} \right),
\end{align*}
where we again invoked Lemma~\ref{lemma:binom-lower} as well as the
fact that $s \leq r/2$. As for the marginal probability in the
denominator, we need to upper bound

\begin{align*}
\Pbb\left[\Ac_1 \mid \uniqint(y_{i^*}, y_{j^*})\right]
&\leq \left(\frac{s}{r} +
\frac{(2s-1)^2\binom{r-2}{s-2}}{\binom{r}{s}}\right)^2 \\
&\leq \left(\frac{s}{r} + \frac{(2s-1)^2(s-1)^2}{(r-1)^2}\right)^2
\leq \frac{s^2}{r^2}\left( 1 + \frac{4s^3}{r}\right)^2,
\end{align*}
since for each $y_i$ and $y_j$, $\Ac_1$ can be satisfied in two ways:
choose the unique element from $\nbd_B(y_{i^*})\cap\nbd_B(y_{j^*})$ or
choose at least two elements from $2s-1$ elements in
$\nbd_B(y_{i^*})\cup\nbd_B(y_{j^*})$.

Using the above two inequalities, we have:
\begin{align*}
  &\Pbb\left[ (y_i, y_j)\in G_{\corr(\rho)} \mid \Delta(y_i, y_{i^*},
  y_{j^*}), \Delta(y_j, y_{i^*}, y_{j^*}), \uniqint(y_{i^*},
  y_{j^*})\right] \\
&\geq \frac{1-\frac{20s^2}{r}} {\left(1+\frac{4s^3}{r} \right)^2}.
\end{align*}
It is easy to verify that $1/(1+x)^2 \leq 1 - x$ for $0 \leq x \leq
(\sqrt{2}-1)/2$. Since $s^3 \leq r/5$, we obtain

\begin{align*}
  &\Pbb\left[ (y_i, y_j)\in G_{\corr(\rho)} \mid \Delta(y_i, y_{i^*},
    y_{j^*}), \Delta(y_j, y_{i^*}, y_{j^*}), \uniqint(y_{i^*},
    y_{j^*})\right]\\
  & \geq \left(1-\frac{20s^2}{r} \right) \left(1 - \frac{4s^3}{r}
  \right)\\
  &\geq 1 - \frac{24 s^3}{r}.
\end{align*}
\eprfof


\subsection{Proof of Proposition~\ref{prop:detect-unique-int}}

Let us start with the case when $\uniqint(y_{i^*}, y_{j^*}) = 1$. For
any pair $(y_i, y_j)$ where $y_i$ and $y_j$ are taken from
$\nbd_{G_{\corr(\rho)}}(y_{i^*}) \cap
\nbd_{G_{\corr(\rho)}}(y_{j^*})$, let $E_{ij}$ be the random variable
which is 1 if $(y_i, y_j) \in G_{\corr(\rho)}$.  Then
Lemma~\ref{lem:detect-good-anchor} guarantees $\Pbb(E_{ij} = 1) \geq 1
- 24s^3/r$.
The size of the set being checked in Procedure~\ref{procedure:uniqueint} is
equal to $\sum_{t} E_{S_t}$.
Hoeffding's inequality guarantees that with probability at least $1 -
2\exp(-2\subsamp\gamma^2)$

\begin{equation*}
  \left| \frac{1}{\subsamp} \sum_{t=1}^{\subsamp} (E_{S_t} - \Pbb(E_{S_t} = 1))
  \right| \leq \gamma.
\end{equation*}
Combining with the lower bound on $\Pbb(E_{ij} = 1)$, we obtain that
with probability at least $1 - 2\exp(-2\subsamp\gamma^2)$,

\begin{equation}
  \sum_{t} E_{S_t} \geq \subsamp \left(1 - 24\frac{s^3}{r} \right) -
  \subsamp \gamma.
  \label{eqn:good-size-lb}
\end{equation}
Using $\gamma \leq 1/64$, we see that this quantity is at least
$62 \subsamp/64$ under the conditions of the lemma, which means that
Procedure~\ref{procedure:uniqueint} returns 1.

Now let us consider the case when $\uniqint(y_{i^*}, y_{j^*}) =
0$. Defining $E_{ij}$ the same way as above, we see that by
Lemma~\ref{lem:detect-bad-anchor}, $\Pbb(E_{ij} = 1) \leq 15/16$. Then,
a similar application of Hoeffding's inequality yields this time

\begin{equation}
  \sum_{t} E_{S_t} \leq \frac{\subsamp}{16} +
  \subsamp \gamma,
  \label{eqn:badd-size-ub}
\end{equation}
which is at most $61\subsamp/64$ for $\gamma \leq 1/64$. Hence
Procedure~\ref{procedure:uniqueint} returns 0 in this case.


\subsection{Proof of Proposition~\ref{prop:accuracy-SVD}}

We now prove Proposition~\ref{prop:accuracy-SVD}. We need a couple of
auxilliary results for the proof. We first restate a theorem from
\cite{Vershynin2010}, which we will heavily use in the sequel.

\begin{theorem}[Restatement of Theorem 5.44 from \cite{Vershynin2010}]
\label{thm:rand-matrix-spectralnorm-vershynin}
Consider a $d \times n$ matrix ${W}$ where each column ${w_i}$ of
${W}$ is an independent random vector with covariance matrix
${\Sigma}$. Suppose further that $\twonorm{w_i} \leq
\sqrt{u}$ a.s. for all $i$. Then for any $t \geq 0$, the following inequality
holds with probability at least $1 - d \exp\left(-ct^2\right)$:
\begin{align*}
  \twonorm{\frac{1}{n} W W^T - \Sigma} \leq
  \max\left(\twonorm{\Sigma}^{1/2}\delta,\delta^2\right) \mbox{ where
  } \delta = t \sqrt{\frac{u}{n}}.
\end{align*}
Here $c>0$ is an absolute numerical constant. In particular, this inequality yields:
\begin{align*}
  \twonorm{W} \leq \twonorm{\Sigma}^\frac{1}{2}
  \sqrt{n} + t \sqrt{u}.
\end{align*}
\end{theorem}


In order to bound the errors made in Algorithm~\ref{algo:init-dict}, we
need some additional notation and auxilliary results. For now, let us
consider a fixed pair of anchor samples $y_{i^*}$ and $y_{j^*}$ such
that $\uniqint(y_{i^*}, y_{j^*})$ is satisfied, and wlog, let
$\nbd_B(y_{i^*})\cap\nbd_B(y_{j^*}) = \{a_1\}$. We define the
following sets of interest

\begin{align}
  \hS &=\nbd_{\corr}(y_{i^*})\cap \nbd_{\corr}(y_{j^*}), \nonumber \\
  S &= \{ y_i \in \hS: \nbd_B(y_{i})\cap \nbd_B(y_{i^*})=
  \nbd_B(y_{i})\cap \nbd_B(y_{j^*})=\{a_1\}\}, \mbox{ and }\\
  \widetilde{S} &= \widehat{S} \setminus S.
  \label{eqn:good-anchor-svd-sets}
\end{align}

For the purposes of understanding the errors in
Algorithm~\ref{algo:init-dict}, it would be helpful to decompose each
vector $y_i \in S$ as

\begin{align}
  \breve{y}_i := y_i - x_{1i} a_1,
  \label{eqn:y-minus-a1}
\end{align}
and accordingly define $\breve{Y}_S$ to be the $d\times |S|$ matrix of
all such vectors in $S$. Intuitively, if all the vectors $\breve{y}$
were 0, then Algorithm~\ref{algo:init-dict} can recover $a_1$ via SVD
in a relatively straightforward manner. We start by controlling the
norm of the vectors $y_i$ and $\breve{y}_i$.

\begin{lemma}
  Under the model~\ref{eqn:dict} and given
  assumptions~\ref{eqn:incoherence},~\ref{eqn:sparsity}
  and~\ref{eqn:sparsitylevel} we have for all $i = 1,2,\ldots, n$

  \begin{equation*}
    \twonorm{y_i} \leq \sqrt{2s}M \quad \mbox{and} \quad
    \twonorm{\breve{y}_i} \leq 2 M \sqrt{s} .
  \end{equation*}
  \label{lemma:twonorms}
\end{lemma}

\bprf

The proof is relatively straightforward consequence of our model and
the assumptions. The model allows us to write

\begin{align*}
  \twonorm{y_i}^2 &= \inner{y_i,y_i} = \sum_{a_p, a_q \in \nbd_B(y_i)}
  x_{pi} x_{qi} \inner{a_p,a_q}\\
  &\leq \sum_{a_p, a_q \in \nbd_B(y_i)}  \abs{x_{pi} x_{qi}}
  \abs{\inner{a_p,a_q}}\\
  &= \sum_{a_p\in \nbd_B(y_i)}  x_{pi}^2 \twonorm{a_p}^2 + \sum_{a_p
    \ne a_q \in \nbd_B(y_i)}  \abs{x_{pi} x_{qi}}
  \abs{\inner{a_p,a_q}}\\
  &\leq M^2\left(s + s^2 \frac{\mu_0}{\sqrt{d}}\right)\\
  &\leq M^2\left(s + \frac{1}{2}\right) \leq \frac{3sM^2}{2}.
\end{align*}

Finally, by triangle inequality we further have that
$\twonorm{\breve{y}_i} \leq \twonorm{y_i} + M$.

\eprf


Given this result, we would next like to control the amount of
contribution the $\breve{y}_i$ directions can have in the SVD step of
Algorithm~\ref{algo:init-dict}. Our next result shows that while these
vectors are not zero, their random support along with the incoherence
of our dictionary elements ensures that these vectors are not strongly
aligned with any one direction. We do so by bounding the spectral norm
of the matrix $\breve{Y}_S$.

\begin{lemma}
  \label{lem:spectralbound-YmoonS}
  With the vectors $\breve{y}_i$ defined in
  Equation~\ref{eqn:y-minus-a1}, we have the following bound with
  probability greater than $1-d\exp\left(-c\alpha^2
  \abs{S}\right)$ for any $\alpha > 0$
\begin{align*}
  \twonorm{\breve{Y}_S} \leq
  M \sqrt{s\abs{S}}\left(\frac{\mu_1}{\sqrt{d}} +
  2\alpha \right),
\end{align*}
where $c$ is a universal constant.
\end{lemma}

\bprf

 In order to prove the lemma, we first calculate the spectral norm of
 the covariance matrix of ${\breve{y_i}}$ and then use Theorem
 \ref{thm:rand-matrix-spectralnorm-vershynin}. Note that from Lemma
 \ref{lemma:twonorms}, we have $\twonorm{\breve{y}_i} \leq
 2 M \sqrt{s}$.  We first bound the spectral norm of the
 covariance matrix of ${\breve{y}_i} \in S$ i.e., we bound
 $\twonorm{\mathbb{E}\left({\breve{y}_i}{\breve{y}_i}^T\right)}$. In
 order to do this, we first fix ${w}\in \mathbb{R}^d$ and calculate:

\begin{align*}
  {w}^T \expec{{\breve{y}_i}{\breve{y}_i}^T} {w}
  = \expec{\left({w}^T {\breve{y}_i}\right)^2}
  = \expec{\left({w}^T {A}{\breve{x}_i}\right)^2}
  = \expec{\left({z}^T {\breve{x}_i}\right)^2},
\end{align*}
where we use the notation ${z} := {A}^T {w}$ and $\breve{x}_i$ is the
same as $x_i$ but with $x_{i1}$ set to 0. We further simplify as

\begin{align*}
  {w}^T \expec{{\breve{y}_i}{\breve{y}_i}^T} {w}
  &\stackrel{}{\leq} \expec{\left(\sum_{p = 1}^r z_p
    \breve{x}_{pi}\right)^2} \\
  &\stackrel{}{=} \expec{\sum_{p =1}^r  z_p^2 \breve{x}_{pi}^2}
  + \expec{\sum_{p\neq q = 1}^r z_p z_q \breve{x}_{pi} \breve{x}_{qi}}
  \\
  &\stackrel{}{\leq} \sum_{p =1}^r z_p^2 \expec{\breve{x}_{pi}^2}
  + \sum_{p\neq q = 1}^r \abs{z_p z_q} \abs{\expec{ \breve{x}_{pi}
      \breve{x}_{qi}}} \\
  &\stackrel{}{\leq} \sum_{p =1}^r z_p^2 M^2 \frac{s}{r}
  + 0,
\end{align*}
where the last inequality uses the fact that the values of
$\E[x_{pi}x_{qi}] = 0$, since both of them are
independent mean zero random variables.

Then we can further simplify the upper bound to obtain

\begin{align*}
  {w}^T \expec{{\breve{y}_i}{\breve{y}_i}^T} {w} &\stackrel{}{\leq}
  \frac{sM^2}{r} \twonorm{{z}}^2 \stackrel{(\zeta)}{\leq}
  \frac{sM^2}{r} \cdot \frac{\mu_1^2 r}{d} = \frac{\mu_1^2 M^2
    s}{d},
\end{align*}
where $(\zeta)$ follows from Assumption $(A3)$, since

\begin{align*}
  \twonorm{{z}} = \twonorm{{A}^T {w}} \leq \twonorm{{A}^T}
  \twonorm{{w}} = \twonorm{{A}{A}^T}^{\frac{1}{2}} \twonorm{{w}} \leq
  \sqrt{\frac{\mu_1^2 r}{d}}.
\end{align*}

Recalling that ${w}$ was an arbitrary unit vector, this immediately
yields a spectral norm bound on the expected covariance
\begin{align*}
  \twonorm{\expec{{\breve{y}_i}{\breve{y}_i}^T}} \leq \frac{\mu_1^2 M^2
    s}{d}.
\end{align*}

We are now in a position to apply
Theorem~\ref{thm:rand-matrix-spectralnorm-vershynin} with the matrix
$W = \breve{Y}_S$ of size $d \times |S|$, where $u = \left(2M\sqrt{s}\right)^2$ and
$t = \alpha \sqrt{|S|}$ for some $\alpha > 0$. Doing so yields the
inequality

\begin{align*}
  \twonorm{{\breve{Y}}_S} &\leq \sqrt{\frac{\mu_1^2 M^2 s}{d}} \cdot
  \sqrt{\abs{S}} + \alpha\sqrt{\abs{S}} \cdot 2 M \sqrt{s}\\
  &\leq M \sqrt{s\abs{S}}\left(\sqrt{\frac{\mu_1^2}{d}} +
  2\alpha \right),
\end{align*}
with probability greater than $1-d\exp\left(-c\alpha^2
\abs{S}\right)$.
\eprf


Finally we are in a position to establish a bound on the accuracy of
the SVD step in Algorithm~\ref{algo:init-dict}. Having bounded the
contribution from from the directions apart from $a_1$ in the previous
lemma, we will now lower bound the contribution of the $a_1$
direction, which will ensure that the largest singular vector is close
to $a_1$.

\bprfof{Proposition \ref{prop:accuracy-SVD}}
Recall the definitions of the sets $S$ and
$\widetilde{S}$~\eqref{eqn:good-anchor-svd-sets}. In order for a
vector $y_i \in \widehat{S}$ to end up in $\widetilde{S}$, the event in
Lemma~\ref{lemma:good-clique} has to fail. Hence, if we define $E_i$
to be the random variable which is 1 if $y_i \in \widetilde{S}$, then
we have from Hoeffding's inequality

\begin{align*}
  \abs{\frac{1}{\widehat{n}} \sum_{i=1}^{\widehat{n}} (E_i - \Pbb[E_i  = 1])} \leq
  \sqrt{\frac{2\log(2/\delta)}{\widehat{n}}},
\end{align*}
with probability at least $1 - \delta/2$. From
Lemma~\ref{lemma:good-clique} we further know that $\Pbb[E_i = 1] \leq
s^3/r$ so that

\begin{equation}
  \abs{\widetilde{S}} \leq \frac{\widehat{n}s^3}{r} + \alpha \widehat{n},
  \label{eqn:bound-Stilde}
\end{equation}
with probability at least $1 - \exp(-2\alpha^2 \widehat{n})$. As a consequence,
the size of $S$ is at least

\begin{equation}
  \abs{S} \geq \widehat{n}(1 - s^3/r - \alpha) \geq 9\widehat{n}/10
\label{eqn:bound-S}
\end{equation}
for $\alpha < 1/20$ by our assumption that $s^3 < r/384$.

In order to understand the singular vector $\hat{a}$, we now write the
matrix $\hL$ as the sum of two matrices $L$ and $\widetilde{L}$ as
follows:

\begin{gather*}
  \hL = L + \widetilde{L}, \mbox{ where, }\\
  L := \sum_{y_i\in S} y_i y_i^T \mbox{ and }
  \widetilde{L} := \sum_{y_i\in \widetilde{S}} y_i y_i^T.
\end{gather*}

Recalling our earlier notation $\breve{y}_i~\eqref{eqn:y-minus-a1}$,
we expand $L$ as follows:

\begin{align*}
  L = \sum_{y_i\in S} y_i y_i^T &= \sum_{i:{y_i}\in S} x_{1i}^2 a_1 {a_1}^T +
  \sum_{i:{y_i}\in S} x_{1i} \left( {a_1}{\breve{y}_i}^T +
      {\breve{y}_i}{a_1}^T \right) + \sum_{i:{y_i}\in S}
      {\breve{y}_i}{\breve{y}_i}^T
\end{align*}
We wish to show that ${a_1}$ is close to the top singular vector of
$\hL$. In order to show this, we bound the spectral norms of the
following matrices: $\sum_{i:{y_i}\in S} x_{1i}\left(
{a_1}{\breve{y}_i}^T + {\breve{y}_i}{a_1}^T \right)$,
$\sum_{i:{y_i}\in S} {\breve{y}_i}{\breve{y}_i}^T$ and
$\widetilde{L}$.

Using Lemma \ref{lem:spectralbound-YmoonS}, we first obtain:

\begin{align}
\twonorm{\sum_{i:{y_i}\in S} x_{1i} {a_1}{\breve{y}_i}^T}
  &\leq \twonorm{{a_1}} \twonorm{{\breve{Y}}_S} \twonorm{{x_{S1}}}
\nonumber\\
  &\leq M\sqrt{s\abs{S}}\left(\sqrt{\frac{\mu_1^2}{d}} +
2\alpha\right)\cdot M\sqrt{\abs{S}} \nonumber
\\
&=  M^2 s\abs{S}\left( \frac{\mu_1}{\sqrt{ds}} +
\frac{2\alpha}{\sqrt{s}}\right)
  \mbox{ and, } \label{eqn:spectralbound-cross}\\
\twonorm{\sum_{i:{y_i}\in S} {\breve{y}_i}{\breve{y}_i}^T}
  &= \twonorm{{\breve{Y}}_S {\breve{Y}}_S^T} \leq
2M^2 s\abs{S}\left({\frac{\mu_1^2 }{d}} +  4\alpha^2
\right).
\label{eqn:spectralbound-YS}
\end{align}

Finally, we have the following bound on the spectral norm of
${\widetilde{L}}$:

\begin{align}\label{eqn:bound-Mtilde}
  \twonorm{{\widetilde{L}}} = \twonorm{\sum_{{y_i}\in \widetilde{S}}
    {y_i} {y_i}^T} \leq \abs{\widetilde{S}} \twonorm{{y_i}}^2 \leq
  \abs{\widetilde{S}} 2sM^2.
\end{align}

Using \eqref{eqn:spectralbound-cross}, \eqref{eqn:spectralbound-YS}
and \eqref{eqn:bound-Mtilde}, we now prove the statement of the lemma.
Denote $\theta = \abs{\iprod{{a_1}}{{\widehat{a}}}}$ and $Z = \frac{1}{\abs{S}}\sum_{i:{y_i}\in S} x_{1i}^2$. On one hand, we
have:
\begin{align*}
  \twonorm{{\widehat{a}}^T{\hL}{\widehat{a}}}
  &\stackrel{}{\leq}  \theta^2 Z \abs{S}
  + 2\twonorm{\sum_{i:{y_i}\in S} x_{1i} {a_1}{\breve{y}_i}^T}
  + \twonorm{\sum_{i:{y_i}\in S} {\breve{y}_i}{\breve{y}_i}^T}
  + \twonorm{{\widetilde{L}}} \\
  &\leq  \theta^2 Z \abs{S}
  +  2 M^2 s\abs{S}\left( \frac{\mu_1}{\sqrt{ds}} +
  \frac{2\alpha}{\sqrt{s}}\right) + 2 M^2 s\abs{S}
  \left({\frac{\mu_1^2}{d}} +  4 \alpha^2\right)
  + \abs{\widetilde{S}} 2s M^2   \\
  &\leq   M^2 \abs{S} \left[\frac{Z}{M^2}\theta^2 +
  8s\left(\frac{\mu_1}{\sqrt{ds}} + \frac{\mu_1^2}{d}
  + \alpha^2 + \frac{\alpha}{\sqrt{s}} +
  \left(\frac{s^3}{r}+\alpha\right)\right)\right],
\end{align*}

where the last step uses the bounds~\eqref{eqn:bound-Stilde}
and~\eqref{eqn:bound-S}. On the other hand, we have

\begin{align*}
  \twonorm{{\widehat{a}}^T{\hL}{\widehat{a}}}
  = \twonorm{{\widehat{L}}} &\stackrel{}{\geq}  Z \abs{S}  \cdot
  \twonorm{{a_1}}^2 - 2\twonorm{\sum_{i:{y_i}\in S} x_{1i}
    {a_1}{\breve{y}_i}^T} - \twonorm{\sum_{i:{y_i}\in S}
    {\breve{y}_i}{\breve{y}_i}^T} - \twonorm{{\widetilde{L}}} \\
  &\geq  Z \abs{S} -  2 M^2 s\abs{S} \left(
  \frac{\mu_1}{\sqrt{ds}} + \frac{\alpha}{\sqrt{s}}\right)
  -  2 M^2 s\abs{S} \left({\frac{\mu_1^2 }{d}} +
  4 \alpha^2\right) - \abs{\widetilde{S}}  2 M^2 s \\
  &\geq   M^2 \abs{S} \left[\frac{Z}{M^2} - 8s\left(\frac{\mu_1}{\sqrt{ds}} +
  \frac{\mu_1^2}{d} + \alpha^2 + \frac{\alpha}{\sqrt{s}} +
  \left(\frac{s^3}{r}+\alpha\right)\right)\right].
\end{align*}
Using the above two inequalities, and the fact that $Z \geq m^2$, we obtain

\begin{align*}
  \theta^2 \geq 1 - \frac{16sM^2}{m^2}\left(\frac{\mu_1}{\sqrt{ds}}+ \frac{\mu_1^2
  }{d} + \frac{s^3}{r}\right) - \frac{16sM^2}{m^2} \left(\alpha^2 +
  \frac{\alpha}{\sqrt{s}}\right).
\end{align*}

Now we observe that since $\twonorm{a_1} = \twonorm{\hat{a}} = 1$, we
have

\begin{align*}
  \twonorm{\hat{a} - a_1}^2 &= 2(1 - \theta) \leq 2(1 - \theta^2),
\end{align*}
for $0 \leq \theta \leq 1$, which completes the proof.

\eprfof


\subsection{Approximate recovery guarantee for Algorithm~\ref{algo:init-dict}}

Building on all our work so far, this section presents the main
guarantee for Algorithm~\ref{algo:init-dict}. So far, we have
established that the sub-procedure in
Algorithm~\ref{procedure:uniqueint} correctly detects good anchor
pairs with high probability. Conditioned on this,
Proposition~\ref{prop:accuracy-SVD} shows that we can recover the
dictionary element in this intersection to a bounded error with high
probability. The next theorem, which puts everything together shows
that in an appropriate number of iterations,
Algorithm~\ref{algo:init-dict} will approximately receover \emph{all}
the dictionary elements with high probability.

%
%
%
%


\begin{lemma}[Number of good anchor pairs]\label{lem:edges-corrgraph}
  Suppose we have $n$ examples. Then, we have:
  \begin{align*}
    \Pbb\left\{\cup_{l \in [r]}\abs{\set{(i,j): \nbd_B(y_{i}) \cap
          \nbd_B(y_{j}) = \set{a_l}}} > \frac{ns}{8r}\right\} \geq 1 -
    r\exp\left(\frac{-ns}{64r}\right).
  \end{align*}
\end{lemma}
\bprf
Fix $l \in [r]$. Define the set $S \subseteq [n]$ as follows:
\begin{align*}
  S := \left\{i : a_l \in \nbd_B(y_{i}) \right\}.
\end{align*}
Since for every $i\in [n]$, the probability of $i \in S$ is $\frac{s}{r}$, using standard
Chernoff bounds, we see that:
\begin{align}\label{eqn:size-S}
  \prob{\abs{S} <  \frac{ns}{2r}} < \exp\left(\frac{-ns}{8r}\right).
\end{align}
Consider any two examples $y_i,y_j \in S$. Then,
\begin{align*}
  \prob{\nbd_B(y_{i}) \cap \nbd_B(y_{j}) = \set{a_l}} \geq 1 - \frac{s^2}{r}.
\end{align*}
Dividing the set $S$ into $\frac{\abs{S}}{2}$ disjoint pairs and using Chernoff bounds, we see that
\begin{align}\label{eqn:size-unique-int1}
  \prob{\abs{\set{(i,j):\nbd_B(y_{i}) \cap \nbd_B(y_{j}) = \set{a_l}}} < \frac{\abs{S}}{4}}
  \leq \exp\left(\frac{-\left(1-\frac{s^2}{r}\right)\abs{S}}{16}\right)
  \leq \exp\left(\frac{-\abs{S}}{32}\right).
\end{align}
Using \eqref{eqn:size-S} and \eqref{eqn:size-unique-int1}, we have:
\begin{align*}
  \prob{\abs{\set{(i,j): \nbd_B(y_{i}) \cap \nbd_B(y_{j}) = \set{a_l}}} > \frac{ns}{8r}}
    \geq 1 - \exp\left(\frac{-ns}{64r}\right).
\end{align*}
Using a union bound over different dictionary elements, we have:
\begin{align*}
  \prob{\abs{\set{(i,j)\middle\vert \nbd_B(y_{i}) \cap \nbd_B(y_{j}) = \set{a_l}}} > \frac{ns}{8r} \; \forall \; l \in [r]}
    \geq 1 - r\exp\left(\frac{-ns}{64r}\right).
\end{align*}
\eprf

\begin{lemma}\label{lem:size-Shat}
  In each iteration of Algorithm~\ref{algo:init-dict}, the size of the set $\widehat{S}$ satisfies:
\begin{align*}
  \abs{\widehat{S}} \geq \frac{ns}{4r},
\end{align*}
with probability greater than $1 - \exp\left(\frac{-ns}{16r}\right)$.
\end{lemma}
\bprf
Since $(y_{i^*},y_{j^*}) \in G_{\corr(\rho)}$, from Lemma~\ref{lemma:corrgraph}, we know that
$\nbd_B(y_{i^*}) \cap \nbd_B(y_{j^*}) \neq \emptyset$. Wlog let $a_1 \in \nbd_B(y_{i^*}) \cap \nbd_B(y_{j^*})$.
Since each sample $y_i$ has probability of at least
\begin{align*}
\frac{s}{r} \cdot \frac{\binom{r-2s+1}{s-1}}{\binom{r-1}{s-1}}
\geq \frac{s}{r} \cdot \left(\frac{r-3s}{r-s}\right)^s
\geq \frac{s}{r} \cdot \left(1-\frac{2s}{r-s}\right)^s
\geq \frac{s}{r} \cdot \left(1-\frac{2s^2}{r-s}\right)
\geq \frac{s}{2r},
\end{align*}
of satisfying $\nbd_B(y_{i}) \cap \nbd_B(y_{i^*}) = \nbd_B(y_{i}) \cap \nbd_B(y_{j^*}) = \set{a_1}$,
using Chernoff bounds, we have:
\begin{align*}
  \prob{\abs{i : \uniqint\left(y_i,y_{i^*}\right) \& \uniqint\left(y_i,y_{j^*}\right)} < \frac{ns}{4r}}
      \leq \exp\left(\frac{-ns}{16r}\right).
\end{align*}
Using Lemma~\ref{lemma:corrgraph} now finishes the proof.
\eprf


\subsection{Auxiliary Results}\label{app:aux}

Below, we establish that the incoherence assumption on the dictionary elements leads to a bound on the RIP constant.

\begin{lemma}\label{lem:RIP-A}
  The $2s$-RIP constant of $A$, $\delta_{2\spindex}$ satisfies
  $\delta_{2\spindex} < \frac{2\mu_0 \spindex}{\sqrt{d}}$.
\end{lemma}
\bprf
Consider a $2\spindex$-sparse unit vector $w \in \R^r$ with $\textrm{Supp}(w)=S$. We have:
\begin{align*}
  \norm{Aw}^2 = \left(\sum_{j \in S} w_j a_j\right)^2
	&= \sum_j w_j^2 \norm{a_j}^2 + \sum_{j,l\in S,j \neq l} w_j w_l \iprod{a_j}{a_l} \\
	&\geq 1 - \sum_{j,l\in S,j \neq l} \abs{w_j w_l} \abs{\iprod{a_j}{a_l}} \\
	&\geq 1 - \sum_{j,l\in S,j \neq l} \abs{w_j w_l} \frac{\mu_0}{\sqrt{d}} \\
	&\geq 1 - \frac{\mu_0}{\sqrt{d}} \onenorm{w}^2 \\
	&\geq 1 - \frac{\mu_0}{\sqrt{d}} 2s \cdot \norm{w}^2 = 1 - \frac{2\mu_0 s}{\sqrt{d}}.
\end{align*}
Similarly, we have:
\begin{align*}
    \norm{Aw}^2 \leq 1 + \frac{2\mu_0 s}{\sqrt{d}}.
\end{align*}
This proves the lemma.
\eprf

\begin{lemma}
  For $r > 2, c > 0$, let $0 \leq x \leq r/(2c+1)$. Then $(1 -
  cx/(r-x))^x \geq \exp(-cx^2/(r-x)) \geq 1 - \frac{2x^2}{r-x}$.
  \label{lemma:binom-lower}
\end{lemma}

\bprf

We start by observing that $x/(r-x)$ is an increasing function of $x$
for $x < r$, so that $x < r/(2c+1)$ implies that $cx/(r-x) <
1/2$. Additionally, we have the following fact for any $\theta > 0$

\begin{equation}
  1 - \theta \leq e^{-\theta} \leq 1 - \theta + \frac{\theta^2}{2}.
  \label{eqn:exp-sandwich}
\end{equation}
The first inequality is a consequence of the convexity of
$e^{-\theta}$ while the second one follows since the second derivative
of $e^{-\theta}$ is at most $1$ when $\theta > 0$. Since we have
$x/(r-x) \leq 1/2$, it is easy to see that

\begin{equation*}
  1 - \frac{cx}{r-x} \geq 1 - 2\frac{cx}{r-x} + 2\frac{c^2x^2}{(r-x)^2}.
\end{equation*}
Now applying the inequalities~\eqref{eqn:exp-sandwich} with $\theta =
2cx/(r-x)$, we obtain

\begin{align*}
  \left(1 - \frac{cx}{r-x} \right)^x &\geq \left(1 - 2\frac{cx}{r-x} +
  2\frac{cx^2}{(r-x)^2} \right)^x\\
  &\geq \left( \exp(-2cx/(r-x)) \right)^x = \exp(-2cx^2/(r-x)\\
  &\geq 1 - \frac{2cx^2}{r-x},
\end{align*}
where the second inequality follows from again
using~\eqref{eqn:exp-sandwich}, this time with $\theta = 2cx^2/(r-x)$.
\eprf

\begin{lemma}\label{lemma:X-covar-matrix}
  We have:
\begin{align*}
  \expec{XX^T} = \frac{s}{r} I_{r\times r},
\end{align*}
where $I_{r\times r}$ is the $r\times r$ identity matrix.
\end{lemma}
\bprf
Let $\Sigma := \expec{XX^T}$. We will first calculate the diagonal elements of $\Sigma$:
\begin{align*}
  \Sigma_{jj} = \expec{x_{ji}^2} = \frac{s}{r}   .
\end{align*}
On the other hand, any off diagonal element can be calculated as follows:
\begin{align*}
  \Sigma_{jk} = \expec{x_{ji}x_{ki}} = \expec{x_{ji}}\expec{x_{ki}} = 0.
\end{align*}
This proves the lemma.
\eprf

\clearpage
\newpage
\bibliographystyle{abbrv}
\bibliography{bib}

\end{document}